\documentclass[lettersize,journal]{IEEEtran}
\usepackage{amsmath,amsfonts,bm}
\usepackage{graphicx}
\usepackage[table]{xcolor}
\usepackage{amssymb}

\usepackage{enumitem}
\usepackage{booktabs} % for table formating
\usepackage{stfloats} 
\usepackage{etoolbox}

\usepackage{flushend}
\usepackage{soul}

\DeclareMathOperator*{\argmax}{arg\,max}

\usepackage{amsthm}

\newtheorem{assumption}{Assumption}

\theoremstyle{definition}

\newtheorem{example}{Example}

\usepackage[export]{adjustbox}
\usepackage{lipsum} 

\begin{document}

\title{Census-Based Population Autonomy For Distributed Robotic Teaming}

%\author{Anonymous Authors}% <-this % stops a space
\author{Tyler M. Paine$^{1,2}$, Anastasia Bizyaeva$^{3}$, and Michael R. Benjamin$^{1}$% <-this % stops a space
\thanks{$^{1}$Department of Mechanical Engineering, Massachusetts Institute of Technology, 
        Cambridge, MA 02139, USA
        {\tt\small tpaine@mit.edu, mikerb@mit.edu}}%
\thanks{$^{2}$Woods Hole Oceanographic Institution
        Woods Hole, MA 02543, USA}%
\thanks{$^{3}$Sibley School of Mechanical and Aerospace Engineering, Cornell University, Ithaca, NY 14853, USA  {\tt\small anastasiab@cornell.edu}}}



\maketitle

\begin{abstract}

Collaborating teams of robots show promise due to their ability to complete missions more efficiently and with improved robustness, attributes that are particularly useful for systems operating in marine environments.
A key issue is how to model, analyze, and design these multi-robot systems to realize the full benefits of collaboration, a challenging task since the domain of multi-robot autonomy encompasses both collective and individual behaviors.
This paper introduces a layered model of multi-robot autonomy that uses the principle of census, or a weighted count of the inputs from neighbors, for collective decision-making about teaming, coupled with multi-objective behavior optimization for individual decision-making about actions.  The census component is expressed as a nonlinear opinion dynamics model and the multi-objective behavior optimization is accomplished using interval programming. 
This model can be reduced to recover foundational algorithms in distributed optimization and control, while the full model enables new types of collective behaviors that are useful in real-world scenarios.  To illustrate these points, a new method for distributed optimization of subgroup allocation is introduced where robots use a gradient descent algorithm to minimize portions of the cost functions that are locally known, while being influenced by the opinion states from neighbors to account for the unobserved costs. With this method the group can collectively use the information contained in the Hessian matrix of the total global cost. 
The utility of this model is experimentally validated with fleets of autonomous surface vehicles in three categorically different experiments: an adaptive sampling scenario, a high value unit protection scenario, and a competitive game of capture the flag. 

\end{abstract}

\begin{IEEEkeywords}
Multi-agent systems, field robotics, opinion dynamics, multi-objective optimization.
\end{IEEEkeywords}

%%%%%%%%%%%%%%%%%%%%%%%%%%%%%%%%%%%%%%%%%%
%% Introduction
\section{Introduction}

Autonomous multi-robot systems have recently captured significant attention due to their promise of improved efficiency and robustness in completing tasks. A key issue is how to model, design, and analyze these multi-robot systems to realize the full benefits of collaboration.  One component of the design is the process of forming teams, or sub-groups within a population, for the purpose of optimizing performance.  Another important consideration is how to design the behavior of individuals within a team to realize a collective team strategy.  In this paper we introduce a new method to address these problems by leveraging two complementary models: nonlinear opinion dynamics for collective decision-making and multi-objective behavior optimization. 

The specific challenges we address are how to autonomously change the composition of the teams and maximize team performance as the mission conditions evolve.  In our approach, individuals have the option to choose among a set of teams to join, and they make the choice using locally known information while relying on a relatively low-bandwidth signal from their neighbors, an opinion state, to provide global context.   The advantage of this approach is that it is distributed, uses less bandwidth, and is scalable to large group sizes.  Once an individual selects to become a member of a team it must determine how to best participate as a new teammate.  The process of determining the optimal trajectory for each member in the team can be framed as a maximization of utility, where the utility functions are determined by desired complex behaviors and can be non-convex.  In our approach, individuals solve for the best trajectory via a search over a set of discrete intervals in the domain of possible reference trajectories.

The hierarchical model introduced in this paper is called census-based population autonomy (CBPA).   The name refers to the mechanism of census \cite{OlfatiSaber2004TACConsensusProblem}, or a weighted count of the inputs from neighbors, used to realize the distributed teaming capability in large groups, or populations.  
At the upper level of the hierarchy, the group uses census to make collective decisions about teaming via a recently developed model of nonlinear opinion dynamics (NOD) \cite{Bizyaeva2023OD_TAC}. 
At the lower, individual level, collaborative behaviors, such as solving the fleet traveling salesman problem (TSP) to find joint path plans, are realized via heterogeneous multi-objective behavior optimization \cite{brooks1986,benjamin2004interval}. 
While the model is intentionally general so that it can be applied to many scenarios, we provide a blueprint for distilling the components to recover well known algorithms for distributed optimization, distributed control, and differential multi-agent games.
Moreover, we detail the design and evaluation of the new features of this approach in three categorically different mission scenarios with groups of up to nine uncrewed surface vehicles (USVs).

The key advantages of the model shown in this paper are:

\begin{itemize}
    \item \textbf{Separation of collective objectives and local objectives.}
    With the inclusion of an opinion state, or preferences for each option in a set of alternatives, agents can separately optimize collective objectives and local actions.  Using the interval programming (IvP) method, local action optimization can be accomplished even when the objective function that maps the domain of actions to utility is not convex. 
    \item \textbf{Distributed optimization of partially observed costs. } Populations can collectively optimize a cost even if each no single agent can directly perceive the entire cost.  This optimization process happens without the need for each agent to reconstruct the global environment through additional communication.   As explained later in Section \ref{sec:analysis}, the global cost function can be decomposed into components that are observed and unobserved by the ego agent. Populations use the nonlinear opinion dynamics model to perform both gradient flow via the Hessian and gradient descent of the cost function, jointly optimizing the global costs.  The method takes advantage of the information in the Hessian but does not require the matrix to be inverted, avoiding the need to perform this challenging task in distributed systems.  

    \item \textbf{Agents do not have to agree on the state of the environment as a prerequisite for making a decision.} There is no need for consensus about environmental parameters.  Information about unobserved states or costs is captured by interpreting the opinions of neighbors, and the NOD formulation guarantees the collective decisions are free from deadlock.

\end{itemize}

\subsection{Application to Marine Robotics}
Multi-robot systems have been shown to be particularly useful for tasks in ocean environments because they can search the vast areas of the oceans more efficiently and are more robust to single-point-of-failures in the harsh marine environment. 
Indeed, examples of successful deployments in marine environments include coverage of lakes and rivers \cite{Karapetyan2018ICRADubins, Gershfeld2023RAL}, ocean-front monitoring \cite{mccammon2021JFR, Leonard2007IEEE}, and the Argo float program \cite{2017Argo}.  In this work we build upon these previous successes by investigating a broader range of scenarios.  In addition to considering the traditional scientific mission of adaptive sampling, we explore different types of scenarios including a competitive game and a scenario that requires the collective defense of a high value unit. We also use larger fleets of up to ten marine robots. Previously reported studies, including \cite{mccammon2021JFR} and \cite{Leonard2007IEEE}, used six or fewer robots. The performance of this new approach to collective autonomy was evaluated using one fleet of ten ClearPath Heron USVs,  and a second fleet of six Sea Robotics SR-Surveyor M1.8.

\subsection{Contributions:}

The contributions of this work include:
\begin{enumerate}
    \item A new framework for modeling and designing coordinated autonomy for multi-robot systems that combines nonlinear opinion dynamics (NOD) and multi-objective behavior optimization.   We highlight the connections to existing methods by showing that this model can be reduced to recover many existing methods for multi-agent distributed control and optimization, as well as multi-agent distributed games.
    \item Analysis of key advantages of the model including its interpretation as a second-order collective optimization process where the Hessian and gradient of the global cost function are used to synthesize the parameters and functions within the opinion dynamics. 
    \item Three different experimental evaluations that use two different fleets of USVs, highlighting the ability of this model to generalize to categorically different scenarios and vehicles.  
\end{enumerate}

This manuscript substantially extends the work by Paine et al. \cite{paine2024ICRAGCID} to the problem of teaming by highlighting connections to existing approaches (Section \ref{sec:generality}), formalizing the design of the input to the NOD process to meet team objectives (Section \ref{sec:analysis}),  and evaluating performance in two different scenarios (Sections \ref{sec:HVU_demo} and \ref{sec:CTF_demo}).

The remainder of the paper is organized as follows:  A \textit{qualitative} review of related work is provided in Section \ref{sec:related_work}.  Mathematical notation is defined Section \ref{sec:math_notation}.  The problem definition is given in Section \ref{sec:prob_definition}.  The census model is introduced in Section \ref{sec:model_def}.  A \textit{quantitative} review of other distributed algorithms and their relationship to this model is provided in Section \ref{sec:generality}.  In Section \ref{sec:analysis} we provide analysis of new optimization techniques made possible by this model, and a study of its performance with respect to communication cost. The design and performance in three multi-vehicle missions are reported in Sections \ref{sec:HVU_demo}-\ref{sec:adaptive_sample_demo}.  Conclusions and future research directions are discussed in Section \ref{sec:conc}.

% Related Work
\section{Related Work}\label{sec:related_work}

One popular view of multi-agent autonomy is that collective action at the group “emerges” as a compilation of actions of individuals \cite{reynolds1987flocks}, \cite{valentini2014self} \cite{Reina2015designpattern}, \cite{Leonard2022PublicGoods}.  The underlying assumption is often that the individuals are not highly sophisticated and must rely on relatively simple algorithms based on only local information since they do not posses the ability to perform extensive planning.  The advantage of focusing on designing simple individual actions to drive group behavior is scalability and robustness \cite{Reina2015designpattern} \cite{Leonard2022PublicGoods}.  The major limitation of these approaches is that they often produce behavior that cannot handle complex combinatorial problems \cite{Luc2009CBBA}, or otherwise may not be optimal due to lack of the necessary communication \cite{Dimidov2016RandomKilobots}.

Another view is that group action can be designed from the top down using centralized algorithms that can achieve near-optimal performance at the cost of reduced flexibility and poor generalization  \cite{Leahy2022ScRATCHes}, \cite{Best2019DeMCTS}, 
\cite{DugganTROLongRange}.  If the environment, goals, and agent states are adequately modeled and globally accessible then decentralized optimization techniques can be used to construct a process to compute the best action for each agent.  
However, in practical applications, no agent has oracle knowledge.
Furthermore, the decision space grows so large a analytical solution is not feasible for NP-hard problems \cite{russell2010artificial}.  However, in many cases even a suboptimal solution is sufficient and even matches the performance of a centralized approach \cite{Leahy2022ScRATCHes}.

In the two previously described perspectives there is a fundamental tradeoff between scalability and optimality.  A third approach uses \textit{distributed methods} to balance this tradeoff by combining the simpler, scalable local algorithms with optimal decentralized approaches \cite{nedic2018distributed}, \cite{Shorinwa2024DistOptSurvey}.  In a distributed algorithm, each agent works toward a global objective by processing its own locally sensed data and exchanging information with nearby agents.
Distributed methods require more complex communication about tasks and plans than bottom up approaches, but do not require every agent to reconstruct the global picture like the top-down decentralized approaches \cite{Luc2009CBBA}. Shorinwa et al. in \cite{Shorinwa2024DistOptSurvey}, a recent review of this topic, categorized these methods by their use of classic linear consensus, a technique first formalized by DeGroot \cite{degroot1974reaching}. Shorinwa et al. defines the simplest approaches as distributed gradient descent methods, where individuals use individual gradient descent of their own locally-known convex cost function, whereas more sophisticated versions that include an additional consensus term to increase the rate of convergence and ensure agreement among all agents\cite{Schwager2009DeAdCovControl}.  These more sophisticated versions are classified as distributed gradient decent with consensus \cite{Shorinwa2024DistOptSurvey}.  In both cases the cost functions must be convex, limiting applicability to complex situations.  Moreover, although linear consensus techniques are widely used to improve performance, their usefulness is limited to the problem formulations that are solved by unanimous agreement.

%limit the solution problem formulations to for some problems the scenario is best addressed by dissensus, or when clusters of agents reach agreement among themselves, but the entire group does not reach a unanimous consensus.   
%result is that all agents arrive to the same state come to the same the process can remain slow and is limited to optimizing convex objective functions \cite{Shorinwa2024DistOptSurvey, nedic2018distributed}. }
% TODO:  Review and make stronger...

In this work we synthesize the bottom-up and top-down approaches, by introducing a distributed model that addresses the limitations of existing distributed methods.  With this model groups make collective decisions through a process of consensus \textit{and dissensus}, or the process where subsets of agents reach of agreement with each other, but not the whole group.  In addition, the objective functions for individual behavior can be non-convex, enabling it to address a broader class of problems.  Furthermore, we show that this model generalizes to many foundational distributed methods. 
The key insight is that group decision-making over a small set of options is achieved through scalable, local interactions between neighbors.  The process of pursuing those options as a team is facilitated by more sophisticated individual behaviors achieved through non-convex optimization.

In Section \ref{sec:generality} we complete a detailed mathematical analysis of how this model unifies many of the existing distributed methods described in this section.

%%%%%%%%%%%%%%%%%%%%%%%%%%%%%%%%%%%%%%%%%%
%% Notation and Definitions
\section{Mathematical Notation}\label{sec:math_notation}
Arbitrary constants are denoted as $\eta \in {\rm I\!R}$ and defined within the context in which they are used. Vectors are in bold, e.g. $\bm{v} = (v_1, \dots, v_n) \in {\rm I\!R}^n$. The $n$-dimensional vector of all ones is $\bm{1}_n$.  Matrices are capitalized with entries in lower case, e.g. $P \in {\rm I\!R}^{n \times n}$ with $p_{ij} \in {\rm I\!R}$. Tensors are capitalized in bold with entries in upper case, e.g. $\bm{P} \in {\rm I\!R}^{n \times n \times n \times n}$ with $\bm{P}^{kk}_{ij} \in {\rm I\!R}$.  Diagonal matrices are generated by the operator $\operatorname{diag}(d_{11}, d_{22}, \hdots, d_{nn})$.  The $\ell^p$ norm is $||\cdot||_p$.

The set of indices in the vector $\bm{v}$ that correspond to the largest elements is $\argmax_{1 \leq i \leq n} v_i$.  The index of the first largest element of that set is 
\begin{equation} \label{eq:def_index_first_min_element}
  i_{v_{max}} = \min [\argmax_{1 \leq i \leq n} \ v_i],
\end{equation}
and the corresponding standard unit vector for that index is defined as $e_{v_{max}}$.

\subsubsection{Graph Representations of Networks}
The bi-directional communication network of robots is represented as an undirected graph $\mathcal{G}(\Omega, \mathcal{E})$ with set of nodes $\Omega$ and set of edges $\mathcal{E}$.  The set of agents connected to the $i^{th}$ agent is $\Xi_i$, and $\mathcal{E}= \{ e \ | \ e_{ij} = 1 \ \forall i,j \in \Xi_i, \}$. The degree, or number of nodes connected to node $i$ , is denoted by $\deg(i)$.  Two nodes are considered connected if there exists a path, or a sequence of edges in the graph $\mathcal{G}$, that connect them.  The graph $\mathcal{G}$ is considered connected is there exists a path between any two nodes in $\mathcal{G}$. \cite{bondy1976graph}

\subsubsection{Matrix Representations of Graphs}
The matrix $A \in {\rm I\!R}^{N_a \times N_a}$ is the unweighted adjacency matrix that corresponds to the $\mathcal{G}$ without self loops with $a_{ij} = 1$ if $e_{ij} \in \mathcal{E}$.  The matrix $L = D - A \in {\rm I\!R}^{N_a \times N_a}$ is the weighted graph Laplacian where $D = \operatorname{diag}
(\operatorname{deg}(1), \ \operatorname{deg}(2), \ \hdots \ \operatorname{deg}(N_a))$. 
For an undirected matrix, $A = A^T$. An undirected graph $\mathcal{G}$ is connected if and only if its adjacency matrix $A$ is irreducible, i.e. cannot be converted to block diagonal form via similarity transformations with permutation matrices. \cite{meyer2023matrix}

\begin{assumption}\label{assumpt:undirected_and_con}
The graph is undirected and connected. %, i.e. every vertex is reachable from every other vertex.
\end{assumption}
\noindent
Two eigenvectors of $A$ are used in the analysis that follows:
\begin{itemize}
    \item $\bm{v}^{*_+}$ is a Perron-Frobenius eigenvector of $A$, i.e. an eigenvector that corresponds to its largest eigenvalue $\lambda^{*_+}$. By Assumption \ref{assumpt:undirected_and_con}, $\lambda^{*_+}$ is unique and $\bm{v}^{*_+}$ can be chosen to have all-positive entries $v^{*_+}_i > 0$ \cite[Theorem 11]{Farina2000PostiveLinearSystems}. 
    
    \item $\bm{v}^{*_-}$ is an eigenvector of $A$ that corresponds to its smallest eigenvalue $\lambda^{*_-}$. 
    The entries in any choice of $\bm{v}^{*_-}$ are mixed-sign \cite[Lemma 1]{bizyaeva2021Patterns}.
\end{itemize}
All eigenvectors referenced in this paper are normalized to unit norm.

%%%%%%%%%%%%%%%%%%%%%%%%%%%%%%%%%%%%%%%%%%%
%% Problem Definition
\section{States and Problem Definition}\label{sec:prob_definition}
\subsection{Individual and Collective State}
The physical state for the $i^{th}$ agent is $\bm{x}_i \in {\rm I\!R}^n$ which includes kinematic states (pose and velocities). % in addition to other states such as fuel level.  
%The state vector $\bm{x}$ importantly does not contain information about the \emph{opinion state} of the agent which is described later in this section. 
%
Each agent also maintains an opinion of $N_o$ options using the following notation adopted from the work of Bizyaeva \textit{et al.} \cite{Bizyaeva2023OD_TAC}:  Opinions of the $i^{th}$ agent are modeled as a vector $\bm{z}_i$, where $z_{ij}$ is the opinion of the $i^{th}$ agent about the $j^{th}$ option, and the opinions of each agent sum to zero.  More formally, $\bm{z}_i \in \bm{1}^{\perp}_{N_o}%\frac{1}{N_o} 
\subset  {\rm I\!R}^{N_o}$. The projection onto $\bm{1}^{\perp}_{N_o}$ is defined as $P_0 = I_{N_o} - \frac{1}{N_o} \bm{1}_{N_o} \bm{1}^T_{N_o}$. Following the convention in \cite{Bizyaeva2023OD_TAC} positive values for $z_{ij}$ imply the $i^{th}$ vehicle has a positive opinion about the $j^{th}$ option.  We assume there are $N_a$ vehicles communicating over a graph $\mathcal{G}$, and define the collective physical state is $\bm{x} = (\bm{x}_1, \dots,\bm{x}_{N_a}) \in {\rm I\!R}^{n \cdot N_a}$ and the collective opinion state is $\bm{z} = (\bm{z}_1, \dots,\bm{z}_{N_a}) \in {\rm I\!R}^{N_o \cdot N_a}$

\begin{figure}[ht]
  \centering
    \includegraphics[width = \columnwidth]{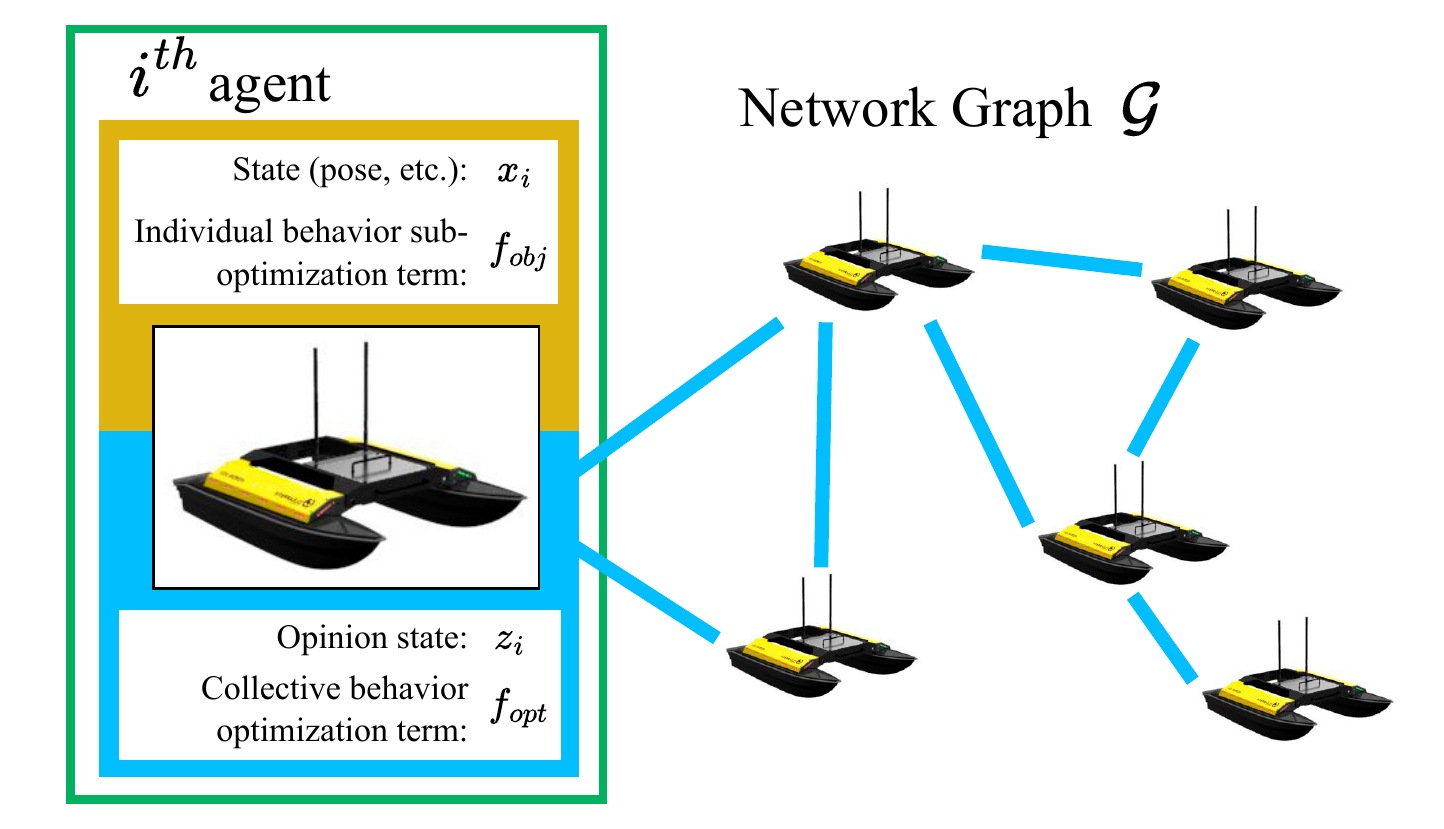}
  \caption{Overview of states: poses $\bm{x}$, opinions $\bm{z}$, and network graph $\mathcal{G}$. } 
  \label{fig:model_overview}
\end{figure}

%\subsection{Network State}

%\subsubsection{Graph}
%The bi-directional communication network of robots is represented as an undirected graph $\mathcal{G}(\Omega, \mathcal{E})$ with set of nodes $\Omega$ and set of edges $\mathcal{E}$.  The graph is assumed to be connected, e.g. Assumption \ref{assumpt:undirected_and_con} is true.  

\subsection{Problem Definition}\label{sec:overal_problem_def}
Consider a scenario where $N_a$ robots are collaborating with an objective that can be expressed as either maximizing a global reward $R_{g}$, or equivalently minimizing a global cost $J_{g}$.  Individual robots in the set are uniquely numbered $1, \ 2, \dots \ N_a$. 
The objective can be decomposed into $N_o \geq 2$ sub-objectives that benefit from the parallelization of tasks, thus incentivizing the formation of teams of robots that tackle these sub-objectives together. In general, $N_o$ can be greater than $N_a$, but in this paper we show examples where $N_o < N_a$.  A team $\mathcal{T}$ is a set of enumerated robots, and these teams are numbered, i.e. the assignments in a two-team, five-robot scenario could have the following description: $\mathcal{T}_1 = \{1,5\}$, $\mathcal{T}_2 = \{2,3,4\}$. 
Any robot can participate in any sub-team.
Individual robots can perceive their local environment, including intrinsic and extrinsic reward signals. 
The state of the environment and capability of each robot changes with time. 
Individuals can communicate to their immediate neighbors on the network graph $\mathcal{G}$, but not the entire group.  

At periodic intervals during the scenario the group and individual robots needs to make the following decisions:
\begin{itemize}
    \item The group must determine the best team assignment for each of the $N_a$ robots that maximizes the (expected) global reward $R_{g}$. 
    \item Each robot must determine the best trajectory to follow, or action that maximizes the (expected) global reward $R_{g}$, while considering the actions of its teammates. 
    \item Each robot must determine the best control input to achieve its own desired trajectory. 
\end{itemize}

\subsubsection*{Remark}
This problem is different from the classic problem of coalition formation where robots are assigned to a set number of coalitions each defined by a task \cite{Vig2006CoalitionFormation, Mazdin2021DistCoalitionForm}. First, the utility, or reward, gained from achieving a task can vary with time.  Moreover, although all robots are assumed to have the same capability or skill, their ability to realize the value from participation in a coalition depends on their state, and the state of the group.  This fact is a departure from the central premise of the classic coalition problem where the capability of a coalition to complete a task is a linear combination (often a simple summation) of the capabilities of its members  \cite{Vig2006CoalitionFormation}.

This problem is also different from the classic problem of distributed optimization and control where robots adjust their state to maximize a local objective \cite{Shorinwa2024DistOptSurvey}.  The details are covered in Section \ref{sec:generality}.

For these reasons, this type of problem is situated between the classic problems of distributed coalition formation and distributed control.

%%%%%%%%%%%%%%%%%%%%%%%%%%%%%%%%%%%%%%%%%%%
%% Background
\section{Background}
\subsection{Nonlinear Opinion Dynamics}
\begin{figure}[ht]
  \centering
    \includegraphics[width = 0.8\columnwidth]{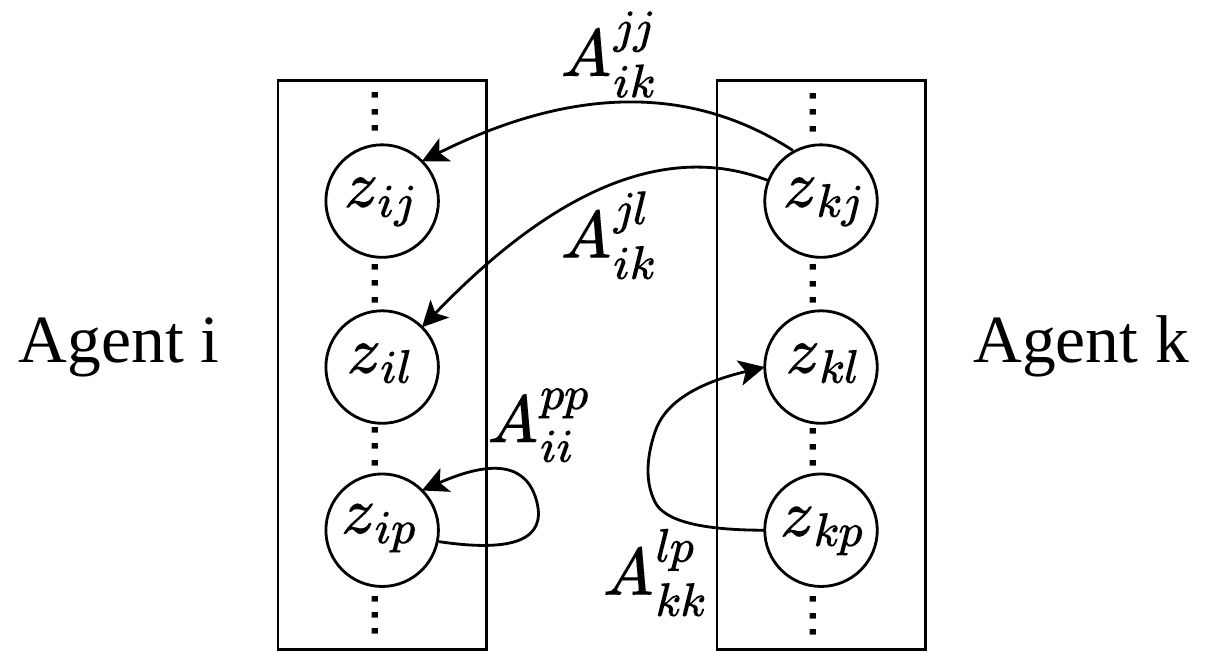}
  \caption{Influence connections in the nonlinear opinion dynamics model (\ref{eq:op_dynam1}).  Adapted from \cite{Bizyaeva2023OD_TAC}. } 
  \label{fig:NOD_influence_dia}
\end{figure}

To model opinion updates of the $N_a$ communicating agents we use the recently  proposed Nonlinear Opinion Dynamics (NOD) model with heterogeneous inter-option coupling from \cite{Bizyaeva2023OD_TAC},
\begin{subequations}\label{eq:op_dynam1}
\begin{align}
\dot{\bm{z}}_i =& P_0 F_i(\bm{z}), \\
F_{ij}(\bm{z}) =& -d_i z_{ij} + u_i \sum_{l = 1}^{N_o} S \bigg(\sum_{k = 1}^{N_a} \bm{A}_{ik}^{jl} z_{kl} \bigg) + b_{ij}(t). %\nonumber%\\
%&+ f_{opt}(\bm{x}_i, \bm{x}_k \in \Xi_i), \nonumber
\end{align}
\end{subequations}
The terms $d_i > 0$ and $u_i$ in  (\ref{eq:op_dynam1}) are tunable parameters that represent the resistance to forming strong opinions and the magnitude of attention to social interactions of agent $i$, respectively. 
In the model, a linear resistance to forming strong opinions competes with a positive feedback from neighbors' opinions and exogenous input.
The adjacency tensor that captures network interactions is denoted as $\bm{A}\in {\rm I\!R}^{N_a \times N_a \times N_o \times N_o}$ with entries $\bm{A}_{ik}^{jl}$ that parameterize the influence from the $k^{th}$ agent's opinion about option $l$ on the $i^{th}$ agent's opinion about option $j$. The term $b_{ij}(t)$ is an exogenous time-varying input to agent $i$ on option $j$.% We will refine the form of this input in Section \ref{sec:model_def}.  
The parameters in this model have the following intuitive interpretations \cite{Bizyaeva2023OD_TAC}:
\begin{itemize}
\item $\bm{A}_{ii}^{jj}$ is the intra-agent, same-option coupling.  We restrict $\bm{A}_{ii}^{jj} \geq 0$, and if $\bm{A}_{ii}^{jj} > 0$ the agent's opinion about this option is self-reinforcing.  
\item $\bm{A}_{ii}^{jl}$ is the intra-agent, different-option coupling.  This parameter is used to encode interplay between agent's own opinions.
\item $\bm{A}_{ik}^{jj}$ is the inter-agent, same-option coupling.
\item $\bm{A}_{ik}^{jl}$ is the inter-agent, different-option coupling.
\end{itemize}
These influence connections are shown in Figure \ref{fig:NOD_influence_dia}. 

In homogeneous systems where $\bm{A}_{ik}^{jl_1} = \bm{A}_{ik}^{jl_2} \ \ \forall \ l_1, l_2 = 1, 2, \hdots N_o \ l_1, l_2 \neq j$ and $\bm{A}_{ik}^{j_1j_1} = \bm{A}_{ik}^{j_2j_2} \ \ \forall \ j_1, j_2 = 1, 2, \hdots N_o$, cooperation and competition between agents are determined by the balance between same-option and inter-option coupling strength,  $\bm{A}_{ik}^{jj}$ and $\bm{A}_{ik}^{jl}$, as: \cite{Bizyaeva2023OD_TAC}
\begin{align}
\bm{A}_{ik}^{jj} - \bm{A}_{ik}^{jl} >& \ 0 \  \rightarrow \ \text{Cooperation,} \\
\bm{A}_{ik}^{jj} - \bm{A}_{ik}^{jl} <& \ 0 \  \rightarrow \ \text{Competition.} 
\end{align}

It is shown in \cite[Appendix B]{Bizyaeva2023OD_TAC} that the set $\mathcal{V} = \{ \bm{z} \ s.t. \ \sum_{j = 1}^{N_o} z_{ij} = 0 \}$ is  forward invariant under the flow of \eqref{eq:op_dynam1}.

In this paper, the input $b_{ij}(t)$ is a designed feedback term for optimizing team composition and is discussed in detail in Section \ref{sec:model_def}.
One of our key contributions in this work is to show how to design the form of $b_{ij}(t)$ for a range of missions described in Sections \ref{sec:HVU_demo}-\ref{sec:adaptive_sample_demo}.

\subsection{Attention and Tunable Sensitivity to Input}
In NOD with no exogenous input, i.e. $b_{ij}(t) = 0$, the attention parameters $u_i$ in \eqref{eq:op_dynam1} control a transition in the network between two distinct regimes. When network attention is low, the linear resistance term in \eqref{eq:op_dynam1} dominates, which results in the stability of a neutral opinion state $\bm{z} = 0$. 
When the attention grows large across the network, the social feedback term dominates and the model \eqref{eq:op_dynam1} undergoes a local bifurcation as the neutral equilibrium loses stability. 
In this opinionated regime, multiple simultaneously stable equilibria $\bm{z} \neq 0$ emerge.
Depending on the structure and signs of social relationships encoded in the adjacency tensor $\mathbf{A}$, these new equilibria correspond to either \textit{consensus} or \textit{dissensus}, i.e. agreement or disagreement,  in the group \cite{Bizyaeva2023OD_TAC}.  
For agents with homogeneous attention and resistance, i.e. with $d_i := d$ and $u_i := u$ for all agents $i$, the critical value of attention for which the bifurcation occurs is $u := u^* = d/\lambda_{max}(\mathbf{A})$ where $\lambda_{max}(\mathbf{A})$ is the largest eigenvalue of the adjacency tensor.

We say the group reached a collective decision when its opinion state $\bm{z}$ has converged to a neighborhood of an opinionated equilibrium of \eqref{eq:op_dynam1} post bifurcation. 
The network attention $u_i$ balances a tradeoff between the sensitivity of a collective decision to exogenous input and its robustness against perturbations \cite{Bizyaeva2023OD_TAC,Leonard2024AnnualReviewFastFlex}. 
For homogeneous agents with $u > u^*$ and small $|u-u^*|$, inputs $b_{ij}$ select among the available multistable equilibria by removing one or more equilibria that are maximally out of alignment with the direction of the input vectors. 
This means when $u$ is near its critical value, the group is ultrasensitive to inputs, and even small inputs can lead to sharp transitions in the collective decision. 
Stronger values of attention therefore make the network more robust against fluctuations in input, at the expense of decreasing the group's overall sensitivity and ability to adapt.  See \cite{Bizyaeva2023OD_TAC} for more details regarding the sensitivity of the NOD to parameters $d$ and $u$.

A dynamic attention feedback mechanism was proposed and explored in \cite{Bizyaeva2023OD_TAC,Leonard2024AnnualReviewFastFlex,bizyaeva2021Cascades,franci2021analysis,cathcart2024SNOD} to simultaneously leverage ultrasensitivity of collective decisions in \eqref{eq:op_dynam1} near bifurcation points and to make the final collective decision robust against perturbations.
Following the convention introduced in these works, let the attention of each agent in \eqref{eq:op_dynam1} be dynamic, evolving according to 
\begin{align}
\tau_u \dot{u}_i = -u_i + S_u\bigg( \frac{1}{N_o} \sum_{k=1}^{N_a} \sum_{l=1}^{N_o} \big( \bar{a}_{ik} z_{kl} \big)^2 \bigg), \label{eq:atten_dynam}
\end{align}
where $\tau_u$ is a time constant of the attention dynamics, $\bar{A} = (\bar{a}_{ik})$ is the adjacency matrix of an attention network, and $S_u$ is a saturation function with $\lim_{x \to 0}S_u(x) = u_{-} < u^*$ and $\lim_{x \to \infty}S_u(x) = u_{+} > u^*$.  
When $u_i(0) < u^*$ for all $i$ and a sufficiently strong exogenous input is introduced to one or more agents in a connected network, feedback between \eqref{eq:op_dynam1} and \eqref{eq:atten_dynam} causes an opinion cascade in which the network embraces a collective decision favored by an input \cite{Bizyaeva2023OD_TAC,franci2021analysis,Leonard2024AnnualReviewFastFlex}.
In this paper we will explore the use of \eqref{eq:op_dynam1} for collective decision-making across multiple applications, both with static attention and with dynamic attention feedback of the form \eqref{eq:atten_dynam}.

Finally, for homogeneous agents with $u > u^*$ and small $|u-u^*|$ the collective decision process is deadlock free \cite{Bizyaeva2023OD_TAC}, meaning the agents will select one of the multistable equilibria.

\subsection{Optimization with Interval Programming}
To model individual decision-making, we use the model for layered behavior-based autonomy \cite{brooks1986} and the Interval Programming (IvP) approach \cite{benjamin2004interval} to solve for the optimal action given one or more objectives.  In the Interval Program formulation, the domain of actions are discrete and the only restriction on the utility functions is that they are piecewise linear.  Thus, the utility functions can be non-convex, enabling this approach to be used in many situations to solve for the best combination of actions, as discussed in Sections \ref{sec:HVU_demo}-\ref{sec:adaptive_sample_demo}. 

As introduced in \cite{Benjamin2010MOOS}, a behavior maps the values of decision variables, such as desired heading and desired speed, to a value of utility. 
The decision space, $\mathcal{S}_m$, for each of $m$ decision variables, $r_m$, is assumed to be finite and uniformly discrete, i.e. $r_m \in \mathcal{S}_m \subset {\rm I\!R}$. 
The $q^{th}$ behavior generates an objective function $f_{q}(r_1, r_2, \hdots r_m): (\mathcal{S}_1 \times \mathcal{S}_2 \times \hdots \times \mathcal{S}_m) \rightarrow {\rm I\!R}$.  
More complicated autonomy can be expressed through a combination of several behaviors, and IvP is used to balance competing objectives. 

The multi-objective optimization problem (\ref{eq:individual_dec}) is traditionally written as as \cite{Benjamin2010MOOS}
\begin{align}
r_1^*, r_2^*, \hdots r_m^* =& \argmax_{r_m \in \mathcal{S}_m \ \forall  m} \sum_{q=1}^{N_{active}} w_{q} f_{q}(r_1, r_2, \hdots r_m)
\end{align}
where the utility function of the $q^{th}$ behavior is weighted by $w_{q} \in {\rm I\!R}$ and $N_{active}$ is the number of active behaviors.  Traditionally, a process known as mode selection \cite{Benjamin2010MOOS}, a type of decision tree process, is used to determine which behaviors are active depending on the state of the mission.  These decision trees are defined at the start of the mission and the structure is static.  A contribution of this paper is using the collective input via the NOD process to activate branches of the decision tree.  More details are provided in Section \ref{sec:model_def}. Finally, the optimal decisions of $r_1^*, r_2^*, \hdots r_m^*$ are included in the reference signal $\bm{r}(t)$ for the robot controller. 

%%%%%%%%%%%%%%%%%%%%%%%%%%%%%%%%%%%%%%%%%%%
%% Model Definitions
\section{Model Definition}\label{sec:model_def}
\begin{figure}[ht]
  \centering
    \includegraphics[width = \columnwidth]{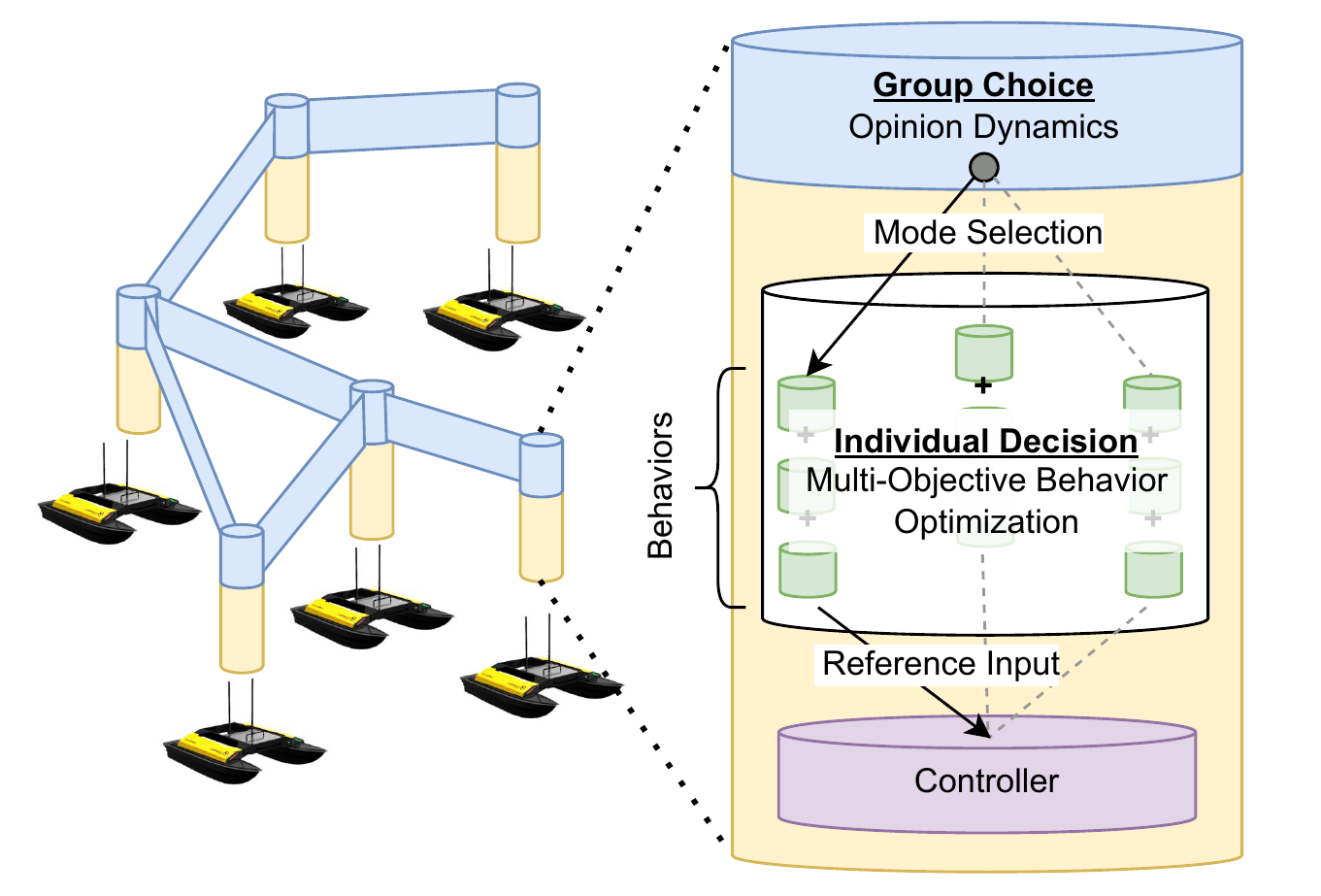}
  \caption{Overview of hierarchical census model adapted from \cite{paine2024ICRAGCID}.  Opinion dynamics is used for distributed team assignment (blue).  Locally, multi-objective behavior optimization is used to determine the best reference input, and a controller is used to execute the action (yellow). } 
  \label{fig:CBPA_overview}
\end{figure}

This section begins with an overview of the hierarchical model structure illustrated in Figure \ref{fig:CBPA_overview}, followed by details of the major components.  

\begin{itemize}[leftmargin=1\parindent]
\item Group choice described in Section \ref{sec:op_dynam}.
\begin{equation}\label{eq:gen_op_dynam}
    \dot{\bm{z}}_i = \underbrace{f_{nod}(\bm{z}_i, \bm{z}_k \in \Xi_i)}_{\text{Nonlinear Opinion Dynamics}} + \underbrace{f_{opt}(\bm{x}_i, \bm{x}_k \in \Xi_i)}_{\substack{\text{Collective Behavior} \\ \text{ Optimization}}} 
\end{equation}
\item Individual decision-making described in Section \ref{sec:ivp}.
\begin{equation}
    \underbrace{r_1^*, r_2^*, \hdots r_m^*}_{\substack{\text{Reference} \\ \text{State}}} = \argmax_{r_m \in S_m \ \forall  m} \underbrace{f_{obj}(\bm{z}_i, \bm{x}_i, \bm{x}_k \in \Xi_i)}_{\substack{\text{Individual Behavior} \\ \text{Sub Optimization}}} \label{eq:individual_dec}
\end{equation}
\item Closed-loop dynamics with a controller designed to compensate for significant disturbances such as wind and waves.  These details are not within the scope of this paper, instead the reader is referred to \cite{mahesh2024safeautonomyuncrewedsurface} as an example implementation.
\begin{equation}
    \dot{\bm{x}}_i = f_{dyn}(\bm{x}_i, \underbrace{\pi(\bm{x}_i, r_1^*, r_2^*, \hdots, r_m^*)}_{\text{Controller}})
\end{equation}
\end{itemize}

\subsection{Group Choice for Teaming via Nonlinear Opinion Dynamics} \label{sec:op_dynam}
In the census-based population autonomy framework, group choice is modeled as a nonlinear dynamical system of opinions \eqref{eq:op_dynam1} where each option has a designed input $b_{ij}(t) = f_{opt}(\bm{x}_i, \bm{x}_k \in \Xi_i)$. 
Since this formulation is completely distributed, the input $f_{opt}(\bm{x}_i, \bm{x}_k \in \Xi_i)$ for each opinion is computed using only locally known information.  This knowledge includes the state of the agent and any opinion information shared between adjacent agents.  The specific form of $f_{opt}$ depends on the mission objectives, and the design of this term in the context of three different scenarios is explained in detail in Sections \ref{sec:HVU_demo}, \ref{sec:CTF_demo}, and \ref{sec:adaptive_sample_demo}.

To connect the group-level choice to agent-level decision-making we use the basis vector $\bm{e}_{z_{max}}(t)$ defined in \eqref{eq:def_index_first_min_element}, which corresponds to the option with the most positive input. 

In both simulation and field experiments we will show that dynamic feedback between opinion dynamics and tunable attention with saturation described in \eqref{eq:atten_dynam} is useful when designing group behavior that must transition from a mode of normal operation to one of emergency which requires a heightened sense of urgency.

\subsection{Individual Decision-Making via IvP}\label{sec:ivp}
In the census based population autonomy framework the individual decision-making is modeled as an optimization problem that is solved by interval programming (IvP) \cite{benjamin2004interval}.  The process computes the optimal solution for the reference trajectory, nominally desired heading and speed, given the objective functions of all active behaviors.  In this framework the state of each behavior, whether it is active or not, is determined by the process of group choice described in Section \ref{sec:op_dynam}.  The remainder of this section provides a description of the multi-objective optimization formulation within the context of the CBPA framework.

In this new framework, the strongest opinion selects which behaviors are active among $N_{b}$ possible behaviors.
\begin{align}
r_1^*, r_2^*, \hdots r_m^* =& \argmax_{r_m \in S_m \ \forall  m} \mathcal{U} \\ 
\mathcal{U} =&\bm{e}_{z_{max}}^T 
A_{c} W
\begin{bmatrix}
f_{1}(r_1, r_2, \hdots r_m) \\
f_{2}(r_1, r_2, \hdots r_m) \\
\vdots \\
f_{N_{active}}(r_1, r_2, \hdots r_m)
\end{bmatrix},
\end{align}
where the diagonal weighting matrix 
\begin{equation}
 W = \operatorname{diag}(w_1 , w_2 ,  \hdots , w_q ). 
\end{equation}
The matrix $A_{c}  \in {\rm I\!R}^{N_O \times N_{b}}$ is the mapping from all $N_{b}$ behaviors to $N_O$ options.  All entries $(A_{c})_{jq} \in \{0,1\}$ are freely chosen design parameters, and could be programmed using logic about the mission such as MODE selection or behavior trees \cite{Benjamin2010MOOS}. 
As a consequence, behavior selection is accomplished, in part, by collective input instead of only using the traditional individual assessment common in the MODE selection or behavior trees.

%%%%%%%%%%%%%%%%%%%%%%%%%%%%%%%%%%%%%%%%%%%
%% Generality
\section{Generality and Connection to Existing Methods}\label{sec:generality} 
\begin{figure}[ht]
  \centering
    \includegraphics[width = \columnwidth]{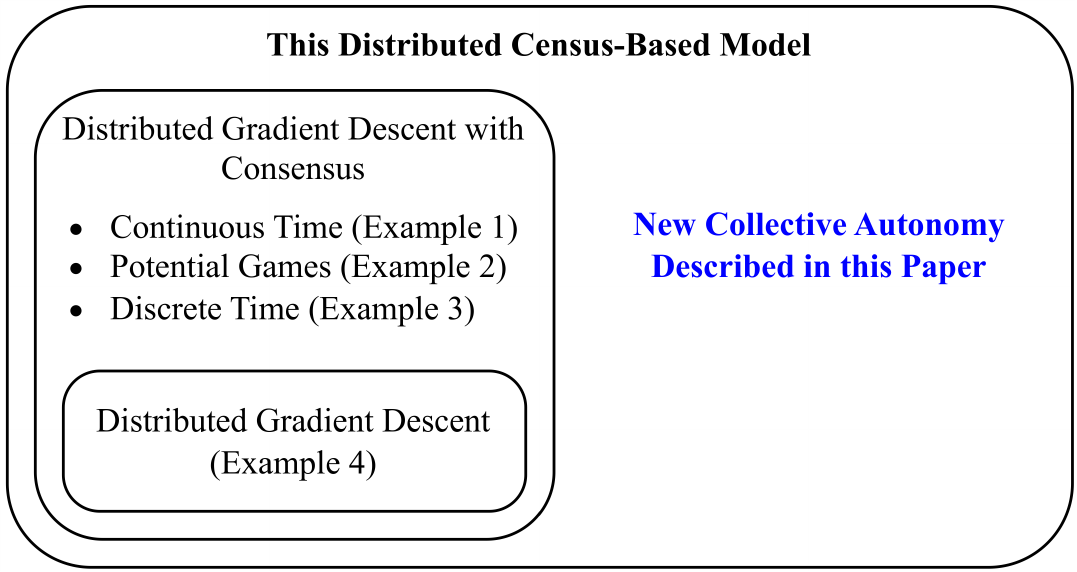}
  \caption{Taxa of decentralized approaches represented by the model in this paper.   The CBPA model can be reduced to recover many existing approaches.} 
  \label{fig:taxa_overview}
\end{figure}

In this section we show how the group choice model is connected to a broad range of distributed algorithms for autonomy \cite{Shorinwa2024DistOptSurvey,marden2009cooperative}.  The model we present is a superset of existing algorithms, as illustrated in Fig. \ref{fig:taxa_overview}. 
By intentionally choosing a few parameters it is possible to limit the range of expression of the full model, reducing it to the classic consensus-based distributed optimization process that is the foundation for many algorithms.

To illustrate this point clearly, consider the reduced $N_a$ agent model for two-option decision-making reported in \cite{Bizyaeva2023OD_TAC}. Due to the projection constraint, for two mutually exclusive options, $z_{i2} = - z_{i1}$ for all agents $i$. Then the opinion formation process for each agent is described by the evolution of a scalar opinion variable $z_{i} := z_{i1}$, whose dynamics are
\begin{equation}
\dot{z}_i = -d_i z_i + u_i \hat{S}\bigg( \alpha_i z_i + \sum_{\substack{k \neq i \\ k=1}}^{N_a} \gamma_{ik} z_k \bigg) + f_{opt}(\bm{x}_i, \bm{x}_k \in \Xi_i) \label{eq:2optionNOD}
\end{equation} 
Near the origin the linearized model in vector form is 
\begin{equation}
\dot{\bm{\delta z}} = -d \bm{\delta z} + u ( \alpha \mathcal{I}_{N_a}+ A \gamma ) \bm{\delta z} + \bm{f}_{opt}(\bm{x}). \label{eq:2optionNOD_vec_lin}
\end{equation}  
We specialize the model to $\alpha_i = 0$, $u_i = 1 \ \forall i$, and $\Gamma \in {\rm I\!R}^{N_a \times N_a}$ is compatible with $A$ with each non-zero entry $\gamma_{ik}$.  

Furthermore, in many of the previously reported algorithms we intend to recover, the opinion state is equivalent to the agent state and the group optimization term is the same as the individual optimization.  Therefore, \textit{only for this generalization} we define the following relationships between states
\begin{align}
\bm{\delta z} = \bm{x} =& \bm{\bar{z}}.  \label{eq:gen_sys_scale_trans} 
\end{align}

With this relationship the dynamics (\ref{eq:2optionNOD_vec_lin}) in vector form are
\begin{align}
\dot{\bm{\bar{z}}} =& (- D  +  \Gamma ) \bm{\bar{z}} + \bm{f}_{opt}(\bm{\bar{z}})  \label{eq:transformed_cons_dynamics1}
\end{align}
where the positive definite diagonal matrix $D = \operatorname{diag}(d_1, \ d_2, \ \hdots \ d_{N_a})$.   In this case the input to agent $i$ is $\bm{f}_{opt}(  \bar{z}_i,   \bar{z}_k \in \Xi_i)$  is a function of the state of the agent and that of its neighbors, and in the following examples this term is typically formulated as a gradient descent of a cost function that influences the state to move towards a optimal solution.   

\begin{example}[Continuous Distributed Gradient Descent with Consensus \cite{Shorinwa2024DistOptSurvey}]\label{ex:cont_dist_opt}
Many problems in distributed optimization and control are of the form:
\begin{equation}
\dot{\bm{\bar{z}}} = \underbrace{- L \bm{\bar{z}}}_{Consensus} - \underbrace{\eta_3 \frac{\partial}{\partial \bm{\bar{z}}}F(\bm{\bar{z}})}_{Optimization}. \label{eq:dist_opt}
\end{equation}
The vector bias $\eta_3 \frac{\partial}{\partial \bm{\bar{z}}}F(\bm{\bar{z}})$ is typically a gradient descent with gain $\eta_3 > 0$  where $F(\bm{\bar{z}})$ is a function to be jointly minimized \cite{Qui2016ContinueousDist} \cite{Pen2017TACDistOpt}.  The reader is referred to \cite{Qui2016ContinueousDist} and \cite{Pen2017TACDistOpt} for more details regarding the properties of $F(\bm{\bar{z}})$ required for convergence.  

The model (\ref{eq:transformed_cons_dynamics1}) can be reduced to the form (\ref{eq:dist_opt}) by setting
\begin{equation}
\gamma_{ij}=\begin{cases}
			a_{ij} > 0, & \text{if $i,j \in \mathcal{E}$}\\
            0, & \text{otherwise}
		 \end{cases} \quad \text{and} \quad d_i = \sum_{k=1, \ k \neq i }^n \gamma_{ij}
\end{equation}
such that $L = D - \Gamma$, completing the generalization. 

One example of a continuous-time control system for multi-robot control is \cite{Schwager2009DeAdCovControl}, where consensus is used to increase the rate of convergence of the estimated field to the true underlying distribution, improving the coverage.

\begin{example}[Potential Games Interpretation]
    When the term $\bm{f}_{opt}(\bar{\bm{z}}) $  can be expressed as a gradient of a scalar objective function, $F(\bar{\bm{z}}) = \sum_{i=1}^{N_a} F_i(\bar{z}_i, \bar{z}_k \in \Xi_i)$ with each $f_{opt}(\bar{z}_i, \bar{z}_k \in \Xi_i) = \frac{\partial F}{\partial \bar{z}_i}(\bar{\mathbf{z}})$ as in Example \ref{ex:cont_dist_opt}, we can interpret \eqref{eq:transformed_cons_dynamics1} as a potential game 
    \cite{monderer1996potential}, which are known to map to broad classes of cooperative control protocols including consensus \cite{marden2009cooperative}.     
    In particular, under the stated assumption, the dynamics \eqref{eq:transformed_cons_dynamics1} admits a global potential function 
    \begin{equation}
        \Phi(\mathbf{z}) = \frac{1}{2} \bar{\mathbf{z}}^T(-D + \Gamma)^T \bar{\mathbf{z}} + F(\bar{\mathbf{z}})
    \end{equation}
    that encodes the global utility. Each agent's local utility is then 
    \begin{equation}
       U_i( \bar{z}_i, \bar{z}_k \in \in \Xi_i) = - \frac{1}{2} d_i \bar{z}_i^2 + \frac{1}{2} \bar{z}_i \sum_{\substack{k \neq i \\ k=1}}^{N_a} \gamma_{ik} \bar{z}_k + F_{i}(\bar{z}_i, \bar{z}_k \in \Xi_i)
    \end{equation}
    and the dynamics in the linearized regime can be interpreted as an exact potential game over a continuous space of actions. 
\end{example}

In most cases the algorithms are implemented and analyzed in discrete time.  We consider this formulation in the next example. 
\end{example}

\begin{example}[Discrete Distributed Gradient Descent with Consensus \cite{Shorinwa2024DistOptSurvey}]\label{ex:disc_dist_opt}
Many applications of distributed optimization, search, and control are implemented in discrete time.  The general form is 
\begin{equation}
\bm{\chi}(t+1) = \underbrace{ (I - \epsilon L)\bm{\chi}(t)}_{Consensus} + \underbrace{\eta_4 F_D(\bm{\chi}(t))}_{\text{Optimization}}
\end{equation}
\cite{nedic2018distributed} where, as explained in \cite{OlfatiSaber2004TACConsensusProblems}, for $d_{max} = \max_{i \in \Omega} L_{ii}$,  $I - \epsilon L$ is a nonnegative and stochastic matrix for all $\epsilon \in (0, 1/d_{max})$. 
Examples include:
\begin{itemize}
	\item \textbf{Distributed learning and control of gradient-climbing swarms} \cite{Choi2009distributed} where consensus is used to match velocities between agents while the optimization term includes several components: a gradient ascent component that directs the agents to the location of the maximum value in the field and other components that balance simultaneous objectives such as collision avoidance.
    \item \textbf{Distributed flocking} \cite{Vasarhelyi2018Flocking}  where consensus is used to match velocities between agents while collision and obstacle avoidance objectives are included in the optimization term.
\end{itemize}
It is straightforward to show the generalization shown in Example \ref{ex:cont_dist_opt} extends to the discrete time case.
\end{example}

\begin{example}[Discrete Distributed Gradient Descent \cite{Shorinwa2024DistOptSurvey}]
The structure in Example \ref{ex:disc_dist_opt} can be further reduced to recover other well-known decentralized algorithms.  The entries in $\Gamma$ are set as $\gamma_{ij} = 0 \ \forall \ i,j$ such that 
\begin{equation}
\bm{\chi}(t+1) = D\bm{\chi}(t) + \underbrace{\eta_4 F_D(\bm{\chi}(t))}_{\text{Optimization}}
\end{equation}

\begin{itemize}
	\item \textbf{Decentralized Monte-Carlo tree search (Dec-MCTS)} \cite{Best2019DeMCTS}: In Dec-MCTS the optimization term is a gradient descent to update the joint probability distribution over all actions for each agent.
    \item \textbf{Particle swarm optimization (PSO)} \cite{Tareq2022PSO}: In PSO the optimization term is a weighted sum of the vector to the best individually observed location and the best locally known location.
\end{itemize}
In general, these decentralized systems that do not include consensus require a fully connected communication graph, or message routing for good performance.  
\end{example}

%%%%%%%%%%%%%%%%%%%%%%%%%%%%%%%%%%%%%%%%%%%
%% Analysis

\section{Second-Order Joint Optimization of Team Assignment with Partially Observed Costs}\label{sec:analysis}
In this section we introduce a new method for distributed optimization made possible by this model and used in multiple scenarios described in this paper.  In later sections we show how to synthesize portions of the model from the mission objectives. 

The objective of a scenario is modeled as a minimization of a joint cost
\begin{equation}
\min_{\bm{z} \in \bar{Z}} f(\bm{z})
\end{equation}
where $f(\bm{z})$ is a function of the option selection of each agent that is determined by an individual's opinion state $z_i$. The following properties of $f(\bm{z})$ are defined as:
\begin{itemize}
    \item $\nabla f(\bm{z}) = \big(\frac{\partial f(\bm{z})}{\partial \bm{z}} \big)^T \in {\rm I\!R}^{N_a}$ a vector where the $i^{th}$ entry is the marginal change in $f(\bm{z})$ with respect to only $z_i$. 
    \item $\frac{\partial^2 f(\bm{z})}{\partial z_i \partial z_j}$ is the marginal change in $f(\bm{z})$ with respect to both $z_i$ and $z_j$. 
\end{itemize}

The domain $\bar{Z}$ is the forward invariant set of \eqref{eq:op_dynam1}. For simplicity in the following development we will define as $\bar{Z} = \{\bm{z} \ |\ ||\bm{z}||_2 \leq 1\}$, which is a reasonable approximation of $\mathcal{V}$ based on the analysis in \cite{paine2024adaptivebiasdissensusnonlinear}. The cost $f(\bm{z})$ is assumed to be separable, meaning it can be decomposed into a sum of locally known objective functions, which is a common assumption in distributed optimization formulations \cite{Shorinwa2024DistOptSurvey}.  In this case we we assume $f(\bm{z})$ can be factored into a sum of two functions:
\begin{align}
\min_{\bm{z} \in \bar{Z}}f(\bm{z}) = \min_{\bm{z} \in \bar{Z}}\sum_{i=1}^{N_a} \big( f_{obs_i}(z_i) + f_{unobs_i}(\bm{z}) \big)
\end{align}
where $f_{obs_i}$ is a function of states observable from the $i^{th}$ agent, while $f_{unobs_i}$ is a function of states that are unobserved.  
In this situation the gradient of the entire function $\nabla f(\bm{z})$ cannot be estimated locally by agent $i$, complicating the use of traditional gradient descent approaches.

Inspired by the large literature on gradient descent methods, as well as the introduction of gradient flow about the Hessian in dynamic games in \cite{hu2023emergent}, we introduce a second order distributed method to solve this problem.  In vector form it is
\begin{align}
\dot{\bm{dz}} = -\eta \big( \nabla f_{obs}(\bm{z}) + H( f(\bm{z}))\bigg|_{z} \bm{dz} \big)
\end{align}
where 
\begin{equation}
    H( f(\bm{z})) = H\big( f_{obs}(\bm{z})\big) + H\big(f_{unobs}(\bm{z})\big)
\end{equation}
is the Hessian of both $f_{obs}$ and $f_{unobs}$ evaluated at $\bm{z}$.  The name follows from the expansion of the gradient up to second order, i.e.  
\begin{equation}
\nabla f(\bm{z} + \bm{dz}) \approx \nabla f(\bm{z}) + H(f(\bm{z})) \bm{dz}.
\end{equation}
Each agent to uses as much local information as possible to calculate the first-order gradient and use the Hessian to capture the dynamics of rewards that are unobserved to an individual agent but are known, collectively, by others. The second-order dynamics are realized by the $i^{th}$ agent's perception of the opinion state $dz$, an essential  component since the value $\nabla f_{unobs}$ is not accessible, limiting the effectiveness of traditional gradient decent methods. Rearranging we have
\begin{align}\label{eq:NOD_2nd_order}
\dot{\bm{dz}} = -\eta H( \bm{f}(\bm{z}))\bigg|_{z} \bm{dz}  - \eta \nabla \bm{f}_{obs}(\bm{z}).  
\end{align}
where, as originally shown in \cite{hu2023emergent} the entries in $H( \bm{f}(\bm{z}))$ are the parameters of influence in the NOD model term $\bm{f}_{NOD}$.  In this way the dynamics of the collective opinion state $\bm{z}$ in the NOD model follows the gradient flow about the Hessian matrix even when individuals cannot observe the entire cost themselves.

This formulation allows for a key insight about one feature of the NOD model, the connection between network structure and input direction.  It is known that the alignment of the input, in this case  $\eta \nabla \bm{f}_{obs}(\bm{z})$,  to the eigenvector $\bm{v}^{*+}$ causes the unfolding of a pitchfork bifurcation and can trigger opinion cascades \cite{bizyaeva2021Cascades}.  In the case of consensus with a symmetric $H( \bm{f}_{unobs}(\bm{z}))$, the vector $\bm{v}^{*+}$ is precisely in the direction of maximum curvature where the gradient of $\bm{f}_{unobs}(\bm{z})$ changes the most.  We can intuitively understand the events of unfolding and cascades as being a consequence of the alignment of the gradient of the observable part of $f$ with the change in gradient of the observable part of $f$.  

The following example highlights the design methodology that is used in experimental missions described later in this paper. 

\begin{example}[Equal Cost Burden Distribution]\label{ex:equal_cost_burden}
This example considers the situation where it is advantageous to balance a group cost, such as battery exhaustion, among agents equally.  Consider the two option scenario where the battery exhaustion rate in steady state is nominally $\xi_1$ ($\xi_2$) for the option $z_i > \ (<) \ 0$.  Let $k_i(z_i)$ be the cost incurred by the $i^{th}$ vehicle, and the cost $\kappa_k$ of other vehicles is not directly observable. An example of a possible $k_i(z_i)$ is the integral of battery consumption over time 
\begin{equation}
   k_i(z_i) = \int_0^t \mathcal{B}_{c} (z_i(t))dt
\end{equation}
where the consumption rate of the option in steady state is 
\begin{equation}
        \mathcal{B}_c(z_i(t)) = \begin{cases} 
           \xi_1, & \text{if $z_i \geq 0$} \\
           \xi_2, & \text{if $z_i < 0$}. 
		 \end{cases}
\end{equation}
In this example the marginal change in cost for selecting the option associated with $z_i \geq 0$ is $\frac{\partial \kappa_o}{\partial z_i} = \xi_1 - \xi_2$. It is possible to use more complex energy consumption models that are beyond the scope of this paper, but the key insight is to use marginal analysis of the estimated cost difference between options. To connect this function to the practical application we describe in Section \ref{sec:HVU_demo}, the marginal change in the steady state battery exhaustion by selecting the option to patrol a large area at high speed instead of loitering in a smaller area at almost zero speed is positive.  

The optimization problem is to minimize the variance
\begin{equation} \label{eq:f_equal_cost_burden}
\min_{\bm{z}\in \bar{Z}} \sum_{i=1}^{N_a} \bigg(\kappa_i(z_i) - \frac{1}{N_a} \sum_{k=1}^{N_a} \kappa_k(z_k) \bigg)^2
\end{equation}
This cost can be factored into a sum of observable and unobserved components. 

\begin{align}
\min_{\bm{z} \in \bar{Z}}& \sum_{i=1}^{N_a} \bigg( \underbrace{\big(1 - \frac{1}{N_a} \big)^2 \kappa_i^2(z_i)}_{f_{obs}} \\ &- \underbrace{2 \frac{1}{N_a} \big(1 - \frac{1}{N_a} \big) \kappa_i(z_i) \sum_{k \neq i}^{N_a} \kappa_k(z_k)  + \frac{1}{N_a^2} \big( \sum_{k \neq i}^{N_a} \kappa_k(z_k)   \big)^2}_{f_{unobs}} \bigg)
\end{align}

The components of the second-order distributed method are
\begin{align}
\nabla \bm{f}_{obs} = 2 \bigg(1 - \frac{1}{N_a} \bigg)^2 \bm{\kappa}(\bm{z}) \frac{\partial \bm{\kappa}}{\partial \bm{z}},
\end{align}
and
\begin{align}
H\big(\bm{f}_{obs}(\bm{z})\big)\bigg|_{z}{}_{ij}= \begin{cases}
    2 \big(1 - \frac{1}{N_a} \big)^2 \frac{\partial^2 (\kappa_i)}{\partial z_i^2 }& \text{if $i=j$}\\
    0& \text{if $i \neq j$},
\end{cases}
\end{align}
and
\begin{align}
H\big(\bm{f}_{unobs}(\bm{z})\big)\bigg|_{z}{}_{ij}= \begin{cases}
    0 & \text{if $i=j$}\\
    \frac{-2}{N_a} \big(1 - \frac{1}{N_a} \big) \frac{\partial (\kappa_i \kappa_j)}{\partial z_i \partial z_j} & \text{if $i \neq j$}
\end{cases}
\end{align}
The term 
\begin{align}
\frac{\partial (\kappa_i \kappa_j)}{\partial z_i \partial z_j} = \begin{cases}
    > 0 & \text{if consensus}\\
    < 0 & \text{if dissensus}
\end{cases}
\end{align}
encodes either the mode of dissensus or consensus. 

For example, in the case of homogeneous agents in consensus the bifurcation is governed by the eigenvalues of $H(\bm{f}_{unobs}(\bm{z}))\bigg|_{z}$, which are $N_a -1$ with multiplicity 1, and $-1$ with multiplicity $N_a -1$.  Thus in the neighborhood of the linearization point the system is stable about the corresponding eigenvector of $H(\bm{f}_{unobs}(\bm{z}))\bigg|_{z}$ that is $\operatorname{span}(\bm{{1}})$, a consensus, which corresponds to the two stability points in the multi-stable NOD model, one with all positive signs, and the other with all negative.   The sign of the solution is governed by the sign of $\bm{1}^T(\nabla \bm{f}_{obs})$ \cite{bizyaeva2021Patterns}.  Thus the desired behavior of jointly minimizing (\ref{eq:f_equal_cost_burden}) is achieved without global knowledge of $\bm{\kappa}$. 

Finally, this simple example shows how including the Hessian term is essential to achieving stability about the entire set that minimizes (\ref{eq:f_equal_cost_burden}), not just $\bm{z} = 0$.  For homogeneous $\kappa_k(z_k)$ is $\bm{z} = \operatorname{span}(\bm{1}) \implies \kappa_i(z_i) = \kappa_j(z_j) \ \forall j \neq i$; with only local knowledge the standard gradient descent method is stable about $\bm{z} = 0$. 

\end{example}

\subsection{Extension to groups with connected network graphs}
In Example \ref{ex:equal_cost_burden}, we assumed the network graph is fully connected.  To extend the formulation to large groups with only connected network graphs (Assumption \ref{assumpt:undirected_and_con}), we state another assumption commonly used in multi-agent reinforcement learning \cite{yang2018mean}.
\begin{assumption}[Mean Field Approximation]
The collective optimization process described by the Second-Order Gradient Flow (\ref{eq:NOD_2nd_order}) can be approximated using only local pair-wise interactions between agents, i.e.
\begin{align}
\dot{\bm{dz}} = \eta \widetilde{H}( \bm{f}_{unobs}(\bm{z}))\bigg|_{z} \bm{dz}  - \eta \nabla \bm{f}_{obs}(\bm{z}).  
\end{align}
where the entries in $\widetilde{H}$
\begin{equation}
    \widetilde{H}_{ij} = \begin{cases}
			H_{ij}, & \text{if $A_{ij} = 1$} \\
           0, & \text{otherwise} 
		 \end{cases}
\end{equation}
\end{assumption}

%%%%%%%%%%%%%%%%%%%%%%%%%%%%%%%%%%%%%%%%%%%
%% Implementation Details
\section{Implementation Details for Experiments}\label{sec:impl_details}

The MOOS middleware was used to facilitate communications in the field work.  MOOS applications and communities utilize the TCP protocol to send messages locally within the vehicles, while UDP is used to send messages to other vehicle communities via the IP address and port of the receiving vehicle.  The information is transmitted over a 2400 MHz RF network using the 802.11b/g/n protocols.

\subsection{USVs}
Two fleets of USVs were used in the experiments. Six Sea Robotics Surveyor M1.8 USVs maintained by the US Military Academic West Point, were used for the game of capture the flag (CTF). In the other two field experiments we used a 16ft OPT-WAM-V USV and a fleet of ten Heron USVs made by Clearpath Robotics.  The Heron USVs involved in the energy balancing scenario had NiMH batteries, which typically have a range of 2-3km on calm water.

All these USVs use the frontseat-backseat paradigm, and the autonomy stack was loaded on the backseat computer. 
The autonomy algorithms described in this paper ran locally on the back seat computer of each vehicle which was a Raspberry Pi 4.

Each vehicle is equipped with a GPS and IMU sensor used for state estimation, and have wireless communication with a router at the shoreside station.   The shoreside station was only used for monitoring, no part of the autonomy was centralized.   

\subsection{Nonlinear Opinion Dynamics}
Each vehicle ran an instance of the Opinion Manager Engine that handled all asynchronous communication of opinion state messages between vehicles.   It was encapsulated within a MOOS App to communicate to other processes using the MOOS middleware. 
%Parameters related to the dynamics of option formation, such as those in (\ref{eq:op_dynam1}) and (\ref{eq:atten_dynam}) are specified as configuration parameters for the Opinion Manager Engine itself.  
The parameters in (\ref{eq:op_dynam1}) were set to $d \in [0.5 \ 1.0]$ and $u > u^*$ for all experiments. 
The iteration and offboard communication frequency of the Opinion Manager Engine were specified to be frequency of 1 Hz. In these experiments each the opinion state messages were approximately 22-25 bytes in size, but the essential components were two real numbers that could be encoded within 4 bytes and retain adequate precision.  Since the bandwidth was relatively low, and the Opinion Manager Engine ignored input from neighbors if no new message was received in the last two seconds,  no significant issues with communication were documented.

\subsection{Interval Programming (IvP)}
Each vehicle solved their individual IvP problem using the MOOS-IvP Helm, which is well documented in \cite{Benjamin2010MOOS} and \cite{benjamin2010UserMan}. The solver ran at 4 Hz, also on the Raspberry Pi 4 computer.

%%%%%%%%%%%%%%%%%%%%%%%%%%%%%%%%%%%%%%%%%%%
%% Exerimentation
%%%%%%%%%%%%%%%%%%%%%%%%%%%%%%%%%%%%%%%%%%%%%%%%%%%%%%%%%%%%%%%%%%%
%\clearpage
\begin{table*}[t]
\fontsize{10.5}{10.5}\selectfont
\centering
\caption{High Value Unit Protection Mission Design (Experiment 1)}
\begin{tabular}{c l l l}
%\toprule
\textbf{Option Structure }($\bm{f}_{nod}$) & \textbf{Option} & \textbf{Option Input} ($\bm{f}_{opt}$)    & \textbf{Behaviors }($\bm{f}_{obj}$) \\
\toprule
\raisebox{-0.8\height}[0pt][0pt]{\includegraphics[height=5\normalbaselineskip]{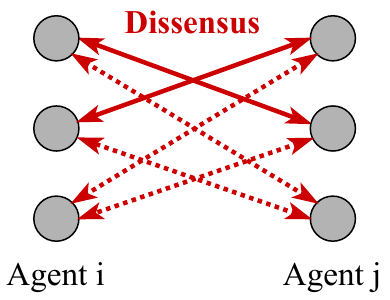}}
  & Patrol & $f_{patrol}$  (Section \ref{sec:balance_batt})  & Voronoi, Voronoi Cell Search \\
  \cmidrule[\lightrulewidth]{2-2} \cmidrule[\lightrulewidth]{3-4}
& Loiter &  $f_{loiter}$  (Section \ref{sec:balance_batt})   & Voronoi \\
  \cmidrule[\lightrulewidth]{2-2} \cmidrule[\lightrulewidth]{3-4}
& Intercept &  $f_{intercept}$ (Section \ref{sec:intercept}) & Trail   \\
  &    &    &  \\
\bottomrule
\end{tabular}
\end{table*}

\section{Experiment 1: High Value Asset Protection Scenario}\label{sec:HVU_demo}
\begin{figure}[ht]
  \centering
\includegraphics[width = \columnwidth]{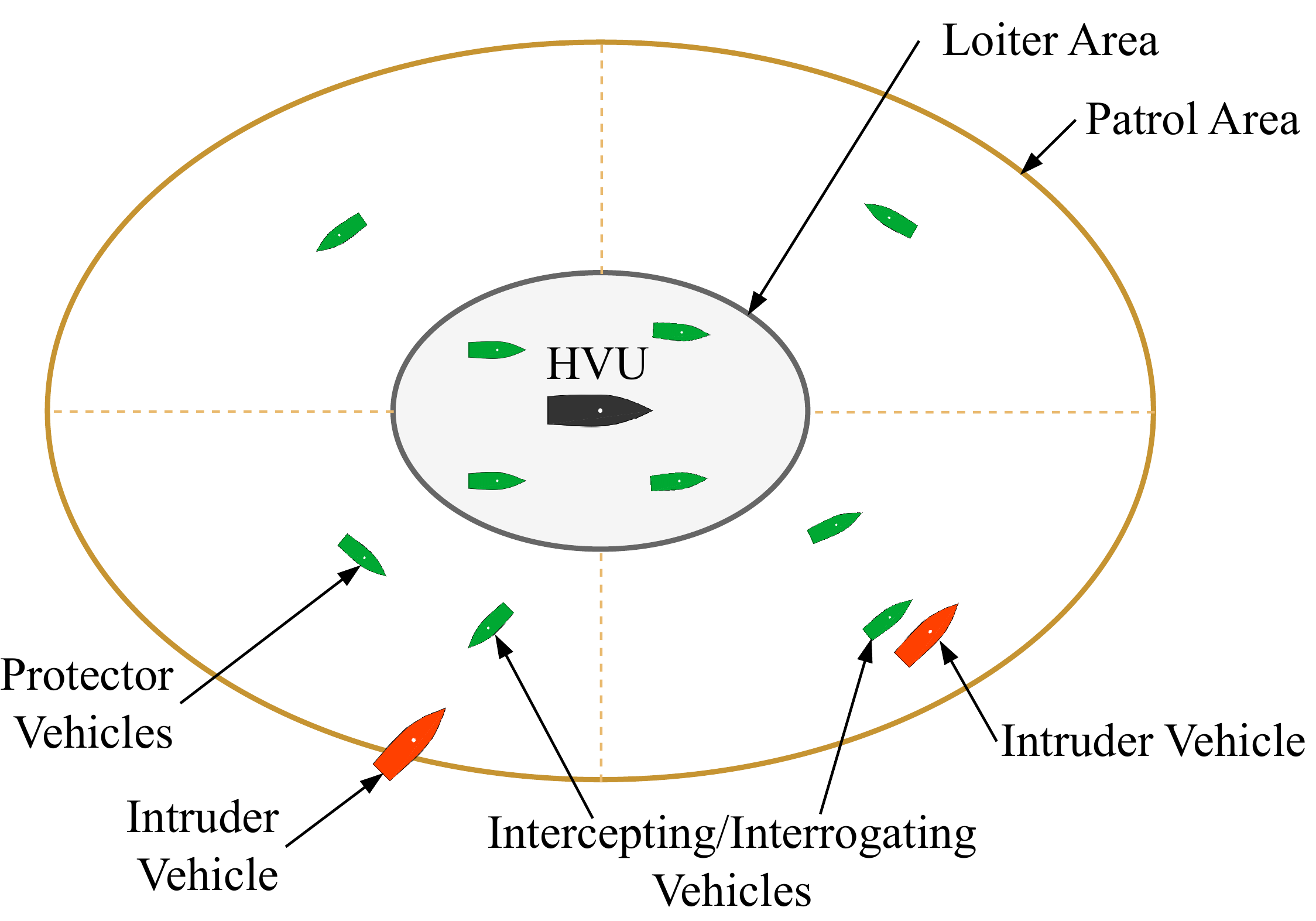}
  \caption{Overview of high value unit (HVU) protection mission.  The moving patrol area is centered on the HVU (in black) with a smaller loiter area closer to the HVU.  Protector vehicles (in green) allocate themselves among the options of patrolling, loitering, or intercepting intruder vehicles (in red).   }
  \label{fig:HVU_mission}
\end{figure}

In this section we consider the scenario where a group of USVs cooperatively protect a high value unit (HVU) by patrolling the surrounding area and interrogating any intruders that approach.  This type of mission is commonly found in naval settings where a ship of high value, such as an aircraft carrier or submarine, is surrounded by a group of ships that are collectively providing protection \cite{OPNAV2020HVU}.  An overview of this mission is shown in Figure \ref{fig:HVU_mission}

We design a system to achieve three objectives at the group level:
\begin{itemize}
    \item Efficiently patrol the area around the HVU.  
    \item Balance the battery level of all USVs by rotating vehicles between patrolling and loitering close to the HVU
    \item Efficiently intercept and interrogate any intruders
\end{itemize}
More details are included in the following sections.

\subsection{Patrolling}\label{sec:patrolling}
Once a USV has chosen the option to patrol, it balances the two objectives of dispersing around the HVU and searching using two behaviors:
\begin{enumerate}
    \item A behavior based on first partitioning a polygon region around the HVU into Voronoi cells and then using Lloyd's algorithm to set a desired trajectory towards the centriod of the cell.  Cells are generated as a function of the location of neighboring agents that are also patrolling, and the utility function is designed to give high utility to values of heading in the direction of the cell centroid. 
    \item An efficient scouting behavior searches within the USV's current Voronoi cell \cite{evans2022practical}.
\end{enumerate}

\begin{figure}[ht]
  \centering
\includegraphics[width = 1\columnwidth]{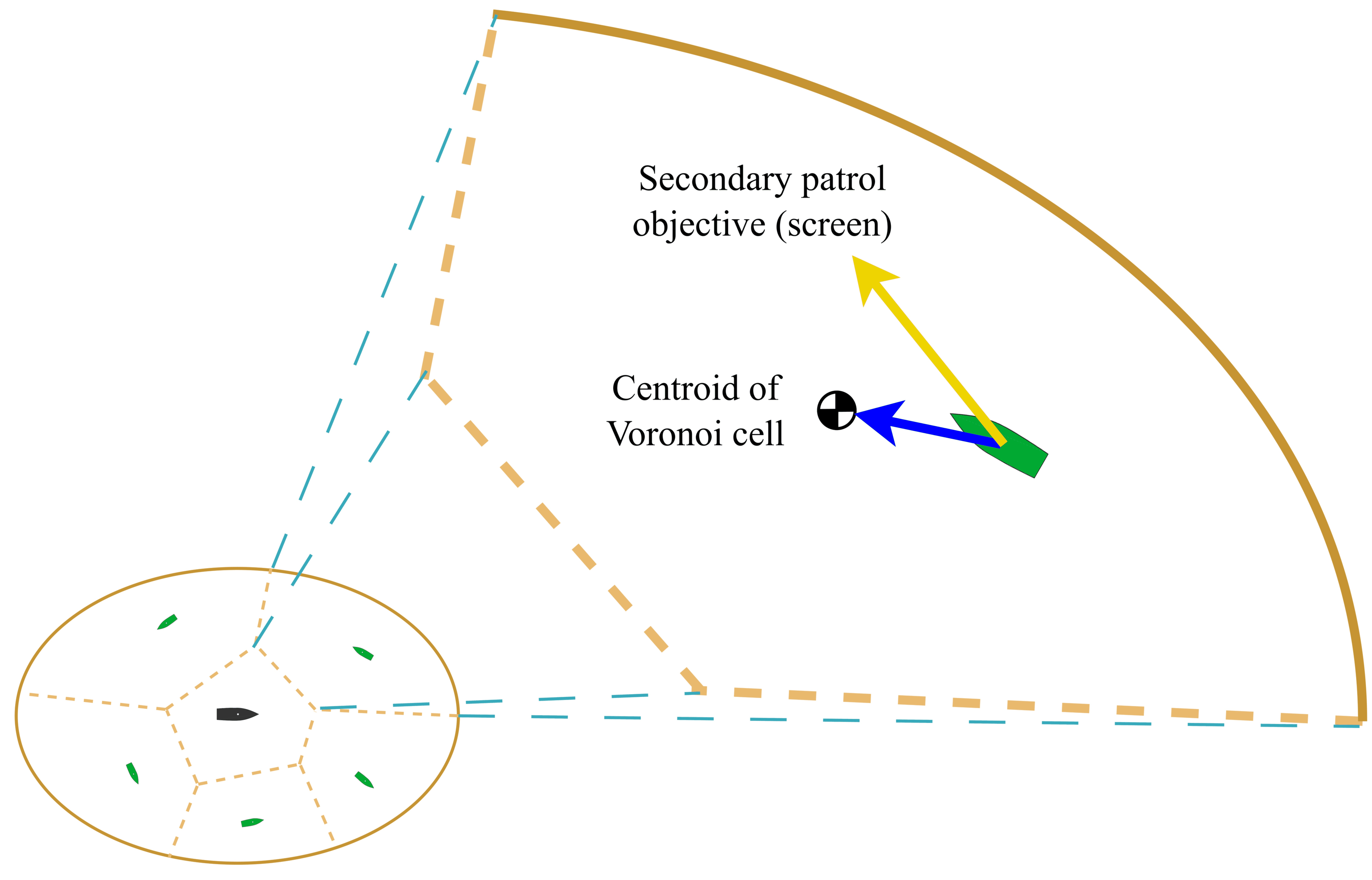}
  \caption{Patrolling vehicle balances two objectives via behavior optimization as investigated in \cite{evans2022practical}. } 
  \label{fig:HVU_patrol_option}
\end{figure}

\subsection{Balancing Battery Level} \label{sec:balance_batt}
Equally distributing the energy cost among protector USVs is primarily accomplished by allocating the vehicles that have the most battery reserves to the task of patrolling which consumes more energy than loitering.   Rotating fresher vehicles to the task of patrol improves the likelihood that the patrol vehicle has the energy to intercept an intruder since they will likely be the optimal choice for the task as described in Section \ref{sec:intercept}.

As introduced in Section \ref{sec:analysis} Example \ref{ex:equal_cost_burden}, the metric to be minimized is the variance of battery exhaustion in the population
\begin{equation}\label{eq:var_batts}
  \mathbb{V}[\bm{\kappa}]  = \sum_{l=1}^{N_a} \bigg(\kappa_l - \frac{1}{N_a} \sum_{k=1}^{N_a}\kappa_k \bigg)^2.
\end{equation}
where $\kappa_i \in [0,  1]$ is the portion battery energy \textit{exhausted} in the $i^{th}$ vehicle.

The goal of minimizing the variance is accomplished via the design of the group choice option inputs $\bm{f}_{patrol}$ and $\bm{f}_{loiter}$. The patrol-loiter allocation is a two-option dissensus case modeled by (\ref{eq:2optionNOD}) where agents with $z_i > 0 \ (< 0)$ opinions choose to patrol (loiter).  The marginal increase in battery exhaustion of selecting to patrol instead of loiter at low speed near the HVU is expressed as
\begin{equation}
    \frac{\partial \kappa_i}{ \partial z_i} > 0.
\end{equation}

We assume that the values of $\kappa_k$ for $k \neq i$ are not known, which is often true in large populations that are not fully connected.  Following the result in Example \ref{ex:equal_cost_burden}, we use the opinion state $z_k$ as a proxy for the $k^{th}$ agent's battery exhaustion and marginal cost of patrolling, given that all vehicles use the same behavior in the IvP layer to generate the guidance law and also use the same low-level controller.
The corresponding option inputs are $f_{patrol} = - \eta_1 \frac{\partial \kappa_i}{\partial z_i} \kappa_i$ and $f_{loiter} = \eta_1 \frac{\partial \kappa_i}{\partial z_i} \kappa_i$ for design constant $\eta_1$

\subsection{Intercepting Intruders} \label{sec:intercept}
When intruders are detected a third option becomes active and the protector USVs use a three-way dissensus to choose whether to allocate themselves into intercepting while still balancing the other two options, patrolling and loitering.  The cost metric $J$ to be minimized is the Euclidean distance all agents in the team have to travel to intercept,
\begin{align} \label{eq:HVU_intercept_cost}
      J =& \sum_{i \in \mathcal{T}_{interpcept}}  J_i \\
    J_i =& \eta_2 \min_{j\in \Upsilon} ||\begin{bmatrix}
        x_i & y_i
    \end{bmatrix}^T- \begin{bmatrix}
        x_{target_j} & y_{target_j}
    \end{bmatrix}^T ||_2,
\end{align}
where $\Upsilon$ is the set of known targets that are broadcast from the HVU and assumed to be known by all protector USVs.  The input $f_{intercept_i}$ is designed as a utility function that is approximately the inverse of (\ref{eq:HVU_intercept_cost}), 
\begin{equation}
    f_{intcpt_i} = \begin{cases}
			f_{intcpt \ max}, & \text{if $J_i < J_{min}$} \\
           f(J_i), & \text{if $J_{min} < J_i \leq J_{max}$} \\
            0, & \text{if $J_i > J_{max}$} ,
		 \end{cases}
\end{equation}
where the linear function 
\begin{equation}
    f(J_i) =f_{intcpt \ max} \bigg(\frac{J_{max} - J_i}{J_{max} - J_{min}}\bigg)  
\end{equation}
In the experiments $f_{intcpt \ max} = 1.0$, $J_{min} = 25 m$, $J_{max}=100m$.
This design uses the relationship between the opinion equilibrium state, the $\bm{v}^{*_-}$, and the input as described in \cite[Corollary IV.1.2]{Bizyaeva2023OD_TAC}.  In summary, the intent is to have the opinion state $\bm{z}$ follow the bifurcation branch $\bm{v}^{*_-} \bigotimes \bm{v}_{ax}$ that corresponds to $(\bm{v}^{*_-})^T \bm{f}_{intercept}$ as shown in \cite[Theorem IV.I]{Bizyaeva2023OD_TAC}. The value of the attention $u$ is sufficiently high that deadlock can not occur. 

\subsection{Experimental Results}
\begin{figure}[ht]
  \centering
\includegraphics[width = 1.0\columnwidth]{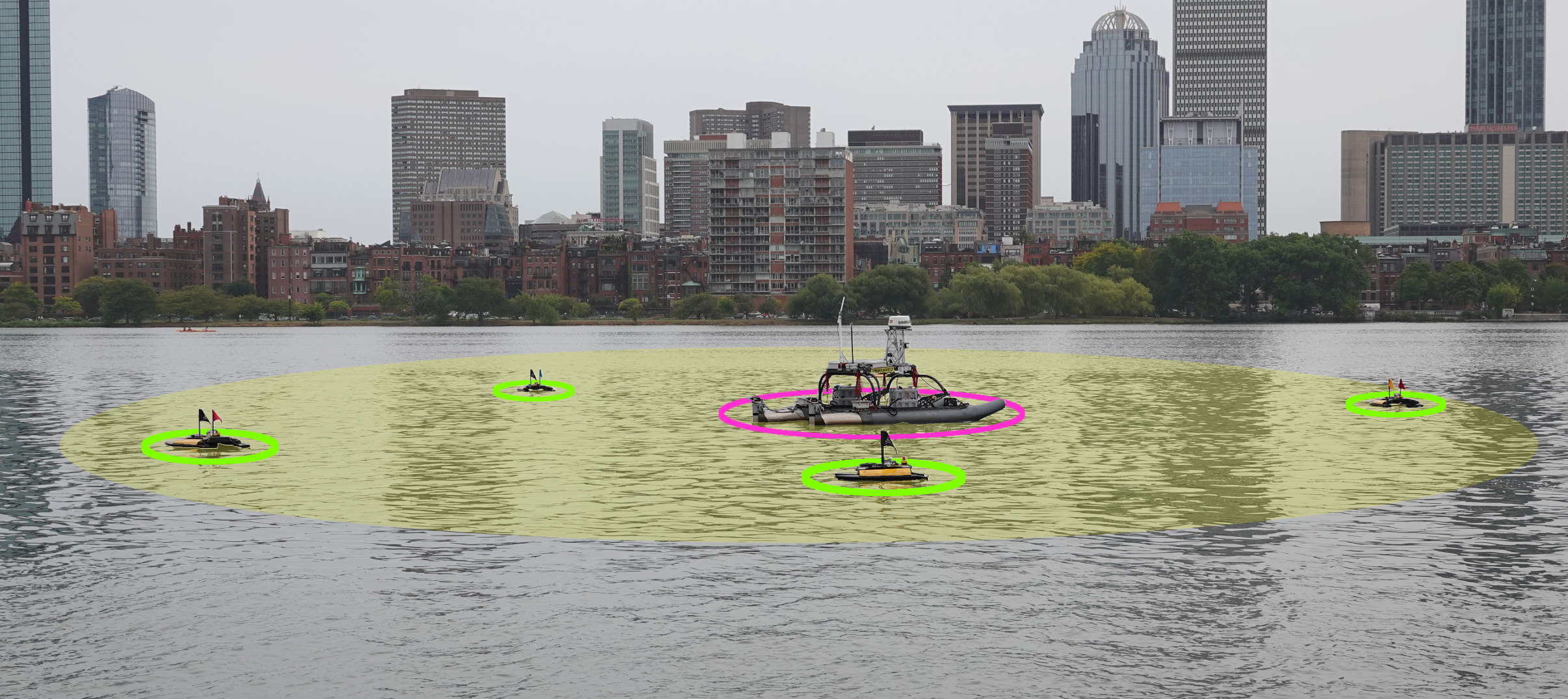}
  \caption{Experimental demonstration of HVU protection mission with a 16ft OPT WAM-V USV as the HVU (circled in pink), and seven Heron USVs (four circled in green in this image) as the protector vehicles.  Demonstration took place on the Charles River near the MIT sailing pavilion.} 
  \label{fig:HVU_demo_image}
\end{figure}

A demonstration of this mission was completed on the Charles River using one 16ft OPT WAM-V USV as the HVU and seven Clearpath Heron USVs as the protector vehicles.  An annotated image of the mission is shown in Figure \ref{fig:HVU_demo_image}.   There were two intruder vehicles: one was another Heron USV following a fixed pattern that periodically violated the patrol region, and one was a human-operated boat that was free to make unscripted intrusions towards the WAM-V based on the operator's judgment.  There were many instances when both intruders were simultaneously inside the patrol region. 

The performance was measured on two metrics: (1) allocating agents to intercept that minimized cost, and (2) reducing variance in battery level.   The allocation performance can be seen in Table \ref{tab:HVU_intercept_results}, where for each of the five times an intruder entered the patrol region the vehicles that selected to intercept where those with the lowest cost per (\ref{eq:HVU_intercept_cost}), matching the performance of a centralized allocation algorithm.   In many cases both intruders where within the patrol region, and the intercepting vehicles implicitly allocated themselves to one of the two targets with the lowest cost to intercept (\ref{eq:HVU_intercept_cost}).

\begin{table}[ht]
\centering
\caption{Intercept cost (Distance in meters) at time of notification of intruder. Allocated vehicles are highlighted in green.}\label{tab:HVU_intercept_results}
\begin{tabular}{| l | c | c | c | c | c |}
\hline
\textbf{Vehicle}  & \multicolumn{5}{|c|}{\textbf{Intrusion Number}}   \\
\hline
& 1 & 2 & 3 & 4 & 5  \\
\hline
Abe & 118.5 & \cellcolor{green!25} 26.3 & 72.3 & 131.3 & 104.5\\
\hline
Ben & \cellcolor{green!25}81.9 & 78.2 & 95.1 & \cellcolor{green!25}95.0 & \cellcolor{green!25}51.0\\
\hline
Cal & 107.9 & 111.6 & 89.6 & \cellcolor{green!25}84.9 & 73.5\\
\hline
Deb & 91.6 & 112.1 & 114.0 & 106.1 & 84.8\\
\hline
Max & 118.4 & \cellcolor{green!25}53.8 & \cellcolor{green!25}43.0 & 106.8 & 96.3\\
\hline
Oak & \cellcolor{green!25}54.7 & \cellcolor{green!25}24.3 & 91.3 & 116.5 & 73.8\\
\hline
Pip & 92.6 & 101.1 & \cellcolor{green!25}23.3 & \cellcolor{green!25}80.7 & \cellcolor{green!25}47.42\\
\hline
\end{tabular}
%\vspace{-5mm}
\end{table}

\begin{figure}[ht]
  \centering
\includegraphics[width = \columnwidth]{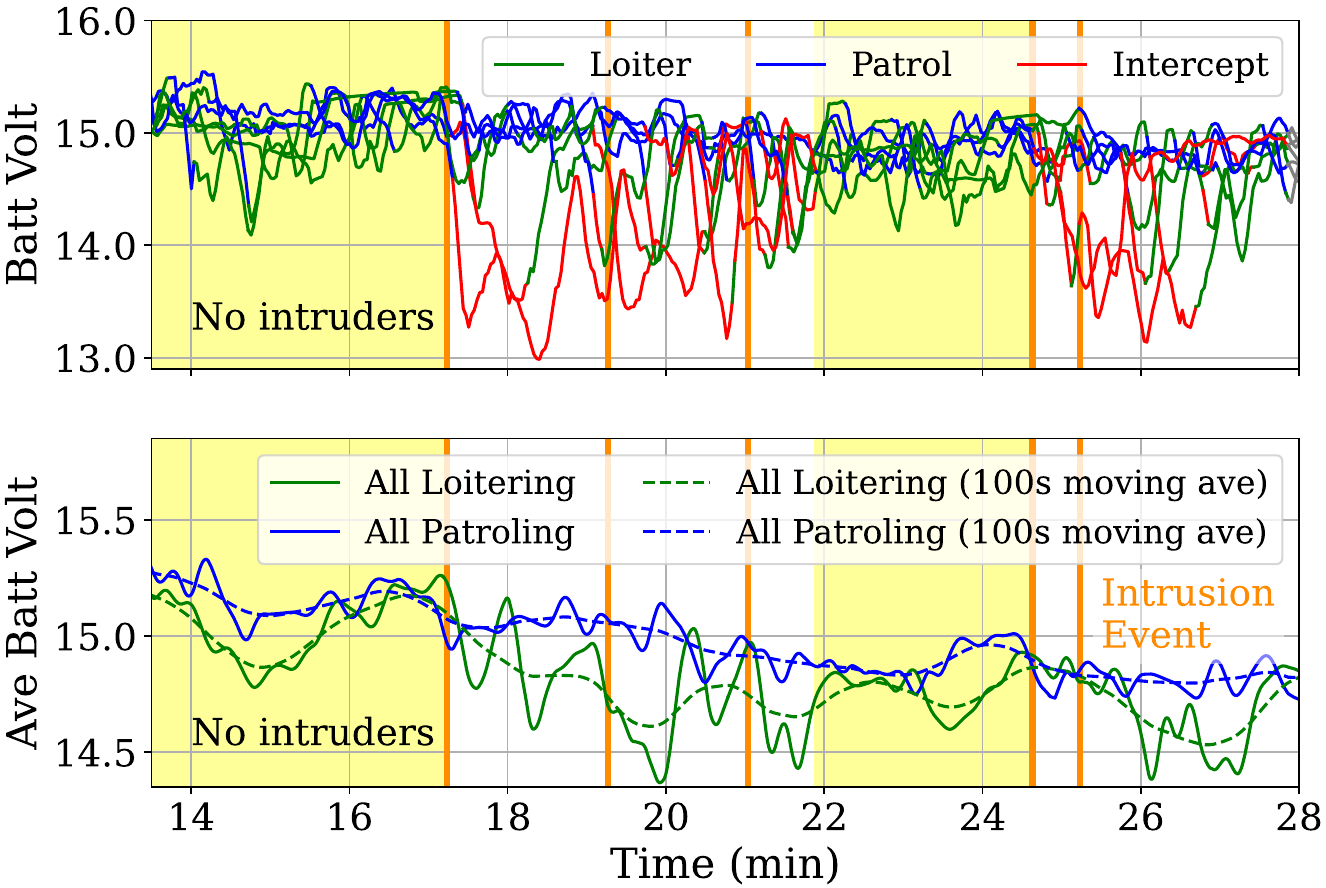}
  \caption{Battery voltage in each Heron during a portion of the HVU mission.  Periods with no intruders are marked in yellow. Battery level traces colored by option the vehicle selected. \textbf{Top} Battery voltage varied during the mission and dropped when vehicles increased speed to chase intruders. \textbf{Bottom} On average vehicles with higher battery levels were allocated to patrol as designed.  } 
  \label{fig:HVU_batt_balance}
\end{figure}

The second metric was the variance in the battery level in the protector Heron USVs.  As shown in the top of Figure \ref{fig:HVU_batt_balance}, when the protector Heron USVs where not intercepting, they switched between patrolling and loitering depending on their battery level.  The bottom plot on Figure \ref{fig:HVU_batt_balance} shows that on average Heron USVs with higher battery levels selected to patrol, the option that consumed battery energy more quickly.  Significant, but temporary, exceptions to this good performance lasted 5-38 seconds and occurred when vehicles were rotating between loiter and patrol, traveling the approximately 20 meters between the two areas at a nominal 1 meter per second  The longer lapses occurred adjacent to an intrusion events and when the group performed an emergency maneuver to move closer to the dock to avoid other river traffic from 23 to 25 minutes into the mission.

\begin{figure}[ht]
  \centering
\includegraphics[width = \columnwidth]{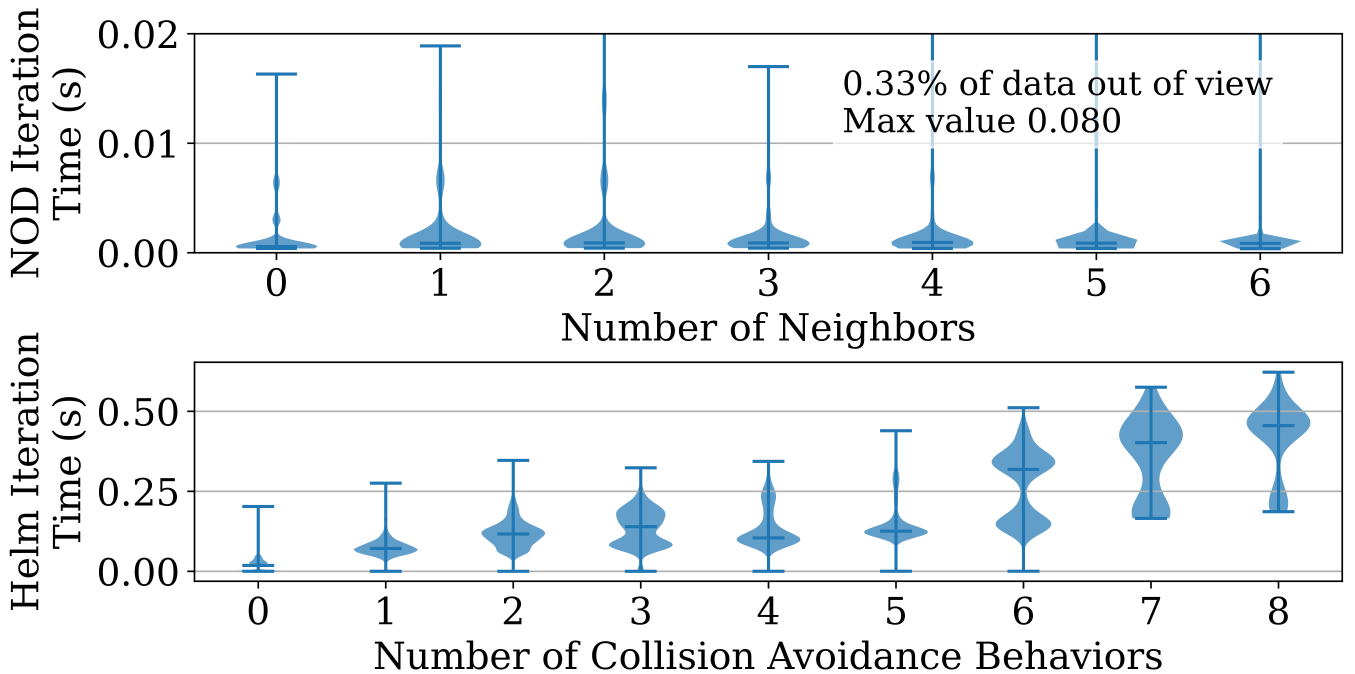}
  \caption{Computation time per iteration by a Raspberry Pi 4 on all Herons during the HVU mission.   \textbf{Top} Average NOD iteration time remains low as neighbors increase \textbf{Bottom} Average Helm iteration remains below 0.25 during the mission where up to five collision avoidance behaviors were active. Iteration increases to 0.5 above five when all the USVs are within the configurable radius of 25 meters that occurred only during deployment or retrieval.} 
  \label{fig:HVU_comp_study}
\end{figure}

Finally, we conducted an analysis of the data to determine how the computational load, quantified by the time to complete an iteration, scales with the number of neighbors and collision avoidance behaviors. The data presented in the top panel of Figure \ref{fig:HVU_comp_study} shows that computational time for the NOD approach does not significantly increase as the number of neighbors increases, even though the theoretical computational cost for the $i^{th}$ agent is $O(N_o \times deg(i))$.  The computational time for the IvP solver, shown in the bottom plot of Figure \ref{fig:HVU_comp_study}, does increase as the number of collision avoidance behaviors increase, and the average time exceeded the 4Hz design rate when six or more where active. However, the times when more than six collision avoidance behaviors were active was rare, only happening at when all vehicles were within a 25 meter radius upon deployment and retrieval, and not during the actual mission experiment when the vehicles were spread out and five or fewer collision avoidance behaviors were typically active. We include this extended data for completeness and as a guide for selecting parameters for collision avoidance behaviors that accounts for expected contact density in future deployments, even though it is not reflective of typical operational conditions.

%%%%%%%%%%%%%%%%%%%%%%%%%%%%%%%%%%%%%%%%%%%%%%%%%%%%%%%%%%%%%%%%%%%%%
%\clearpage
\begin{table*}[t]
\fontsize{10.5}{10.5}\selectfont
\centering
\caption{Capture-The-Flag Game Mission Design (Experiment 2)}
\begin{tabular}{c l l l}
%\toprule
   \textbf{Option Structure }($\bm{f}_{nod}$) & \textbf{Option} & \textbf{Option Input} ($\bm{f}_{opt}$)    & \textbf{Behaviors }($\bm{f}_{obj}$) \\

\toprule
\raisebox{-0.75\height}[0pt][0pt]{\includegraphics[height=3.5\normalbaselineskip]{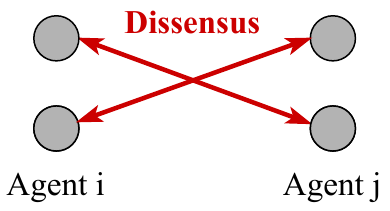}}
  & $f_{attack}$  (Section \ref{sec:attack_vs_defend}) & Attack  & Waypoint, Collision Avoidance \\
  \cmidrule[\lightrulewidth]{2-2} \cmidrule[\lightrulewidth]{3-4}
  & $f_{defend}$  (Section \ref{sec:attack_vs_defend}) & Defend & Trail, Loiter \\
  &    &    &  \\
\bottomrule
\end{tabular}
\vspace{-5mm}
\end{table*}
\section{Experiment 2: Game of Capture-the-Flag }\label{sec:CTF_demo}
\begin{figure}[ht]
  \centering
\includegraphics[width = \columnwidth]{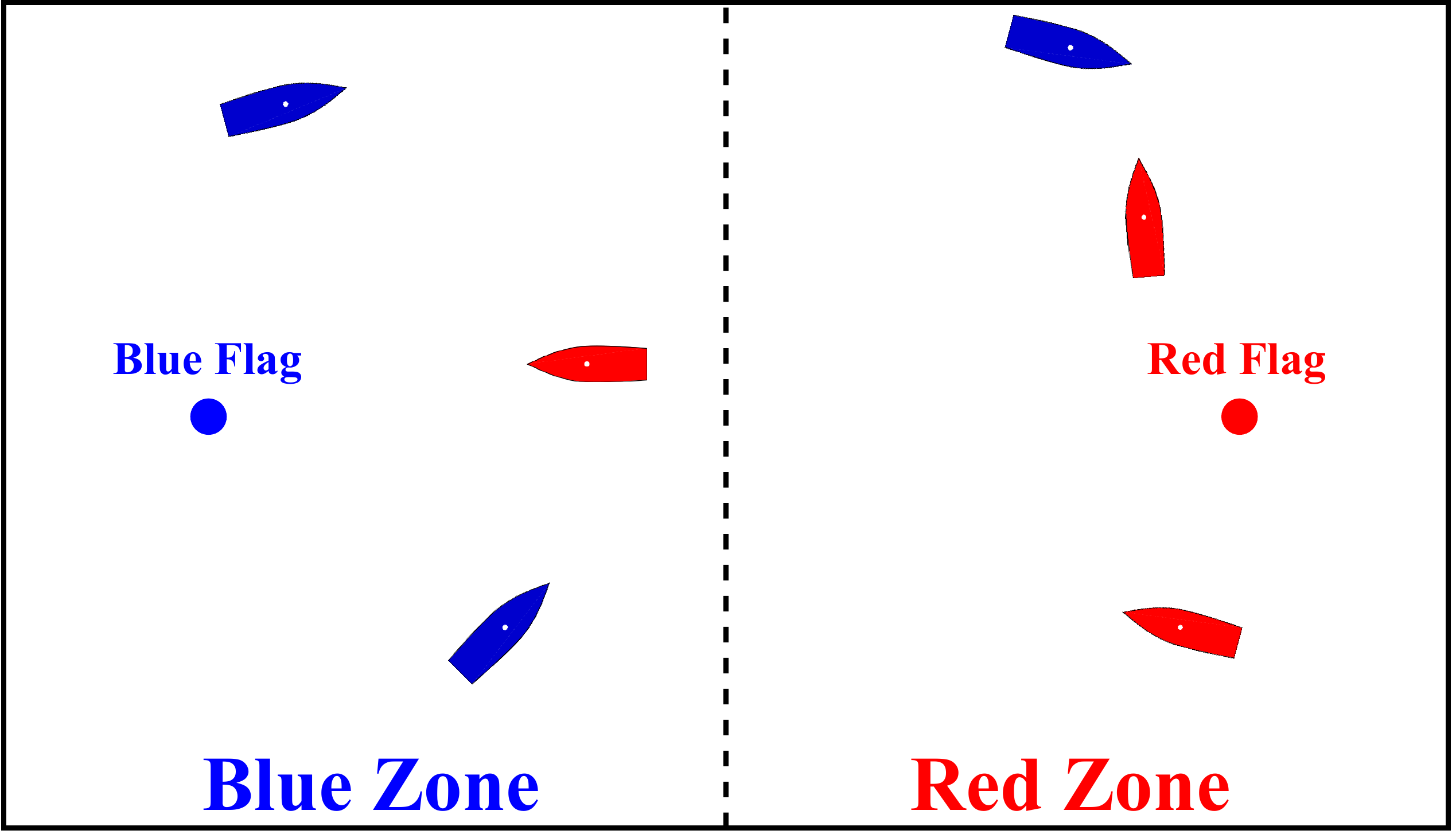}
  \caption{Overview of Aquatics game of capture-the-flag (CTF) \cite{beason2024evaluatingcollaborativeautonomyopposed} \cite{Novitzky2019Aquaticus}.  Two teams (red and blue) of three vehicles score by reaching the opposing team's flag and returning it to their flag location.  Vehicles defending their own team's flag can ``tag out'' intruders once the intruder crosses onto their zone of the field. } 
  \label{fig:CTF_mission}   
\end{figure}

In this section we consider a competitive game of capture the flag (CTF) with two teams, each with three surface vehicles.  This game is commonly known in the literature as Aquaticus \cite{beason2024evaluatingcollaborativeautonomyopposed}, \cite{Novitzky2019Aquaticus}, and a summary overview is shown in Figure \ref{fig:CTF_mission}.   The goal of the game is to score more points than the opposing team, where one point is awarded for a flag grab, and two points for a flag capture, or carrying a grabbed enemy flag back to the team's own base. 

The system balances the objectives of maximizing own rewards and minimizing opponent's rewards.  The key features of our approach are:
\begin{itemize}
    \item Dynamic group allocation of vehicles between the two options of attack and defend. 
    \item Design of multi-objective behaviors that are  heterogeneous.
\end{itemize}

\subsection{Attack vs. defend allocation}\label{sec:attack_vs_defend}
A key advantage of this method over others is the flexibility for vehicles to rapidly change between options of attacking vs defending. This group choice is modeled as a two option dissensus process described earlier in (\ref{eq:2optionNOD}) where positive $z_i >0 \ (< 0)$ implies the vehicle selects the option to attack (defend).
Since the network of three agents is fully connected in this scenario, the smallest eigenvalue $\lambda^{*-}=-1$ has geometric multiplicity of two, and thus the opinion equilibria exist in a 2 dimensional subspace.  Guided by the analysis introduced in \cite{paine2024adaptivebiasdissensusnonlinear}, the inputs $f_{attack}$ and $f_{defend}$ are heuristically designed using the following empirical method based on theoretical properties of nonlinear opinion dynamics. 

In the absence of any input $\bm{f}$ the opinion equilibria exist on a surface that can be approximated as the linear subspace $V$ spanned by the two eigenvectors, i.e. 
\begin{equation}
   V = \operatorname{span}\Bigg\{ \begin{bmatrix}
        -1 \\
        1 \\ 
        0
    \end{bmatrix}, \begin{bmatrix}
        -1 \\
        0  \\
        1
    \end{bmatrix} \Bigg\}
\end{equation}
The equilibria exist in a forward invariant set \cite[Appendix B]{Bizyaeva2023OD_TAC} that lies in $V$.

First, assume all vehicles have identical inputs $f_{attack}$ and $f_{defend}$, which can be expressed as the vector $f_{ave}\bm{1}$ for $f_{ave} = \frac{1}{2}(f_{attack} - f_{defend}) \in {\rm I\!R}$.  The vector $\bm{1}$ is orthogonal to the subspace $V$.  With values of $f_{ave} > 0 \ (<0)$ the surface that contains the equilibria moves parallel to $V$ in the direction of $\bm{1}$ $(-\bm{1})$.  The topology of the forward invariant set changes, however the equilibria exist on a surface that remains approximately parallel to $V$.

The key insight from the design perspective is that values for $f_{ave}$ can be chosen such that the forward invariant set lies in a quadrants that are only permutations of $(+,-,-)$, which means two vehicles are defending, or chosen such that the set lies in the $(-,-,-)$ quadrant, which means all three are defending.  

Second, we relax the assumption of identical inputs, and let $\{\bm{f}_{net} \in {\rm I\!R}^3 : ||\bm{f}_{net}||_2 = \bm{f}_{ave}\}$. Although the topology of the invariant set changes with the direction of $\bm{f}_{net}$, the surface that contains the equilibria remains well approximated by $\bm{f}_{ave}$ as explained in the previous paragraph.  Together a simple heuristic was designed to take advantage of these two insights: 1) the input magnitude, $\bm{f}_{ave}$, was a function of the number of attackers on the own side of the field, and controlled the ratio of attacking vs defending agents, while 2) the direction of $\bm{f}_{net}$, a function of the distance to own flag, was used to control which agents selected to attack vs defend. The result was a strategy flexible and responsive to changes in the state of the game.

\begin{table*}[ht]
\centering
\caption{Game performance in simulation and field competitions }
\begin{tabular}{ l c  c c c c c c c }\label{tab:CTF_comp_results}
\textbf{Ours vs.}  & \textbf{Comp.} &  &  &  & \textbf{Flag} & \textbf{Flag Grabs} & \textbf{Flag} & \textbf{Flag Captures} \\
\textbf{Opponent:}  & \textbf{Type } & \textbf{Games} & \textbf{Win \%} & \textbf{Tie \%} & \textbf{Grabs} & \textbf{Allowed} & \textbf{Captures} & \textbf{Allowed} \\
\hline
Default \cite{beason2024evaluatingcollaborativeautonomyopposed} & Simulated & 200 & 89.0\% & 6.0\% &  3.1 (mean) &  0.95 (mean) & 1.9 (mean) & 0.2 (mean)\\
\hline
Best Rule-Based \cite{beason2024evaluatingcollaborativeautonomyopposed} & Simulated & 200 & 65.0\%  & 11.5\% & 1.9 (mean) & 1.0 (mean)& 0.95 (mean)& 0.35 (mean)\\
\hline
No NOD (ablation) & Simulated & 200 & 50.6\%  & 11.7\% &1.9 (mean) & 2.0 (mean) & 0.8 (mean) & 0.5 (mean)\\
\hline
FS1*  & Field & 1 & N/A & N/A & 3 & 0 & 2 & 0\\
\hline
FS2* & Field & 1 & N/A & N/A & 6 & 0 & 2 & 0 \\
\hline
\multicolumn{5}{l}{*Entry in 2024 AAMAS Maritime Capture the Flag Competition \cite{AAMAS2024CTFComp}}
\end{tabular}
\vspace{-5mm}
\end{table*}

\subsection{Multi-objective behavior design}\label{sec:ctf_behavior}
\begin{figure}[ht]
  \centering
\includegraphics[width = \columnwidth]{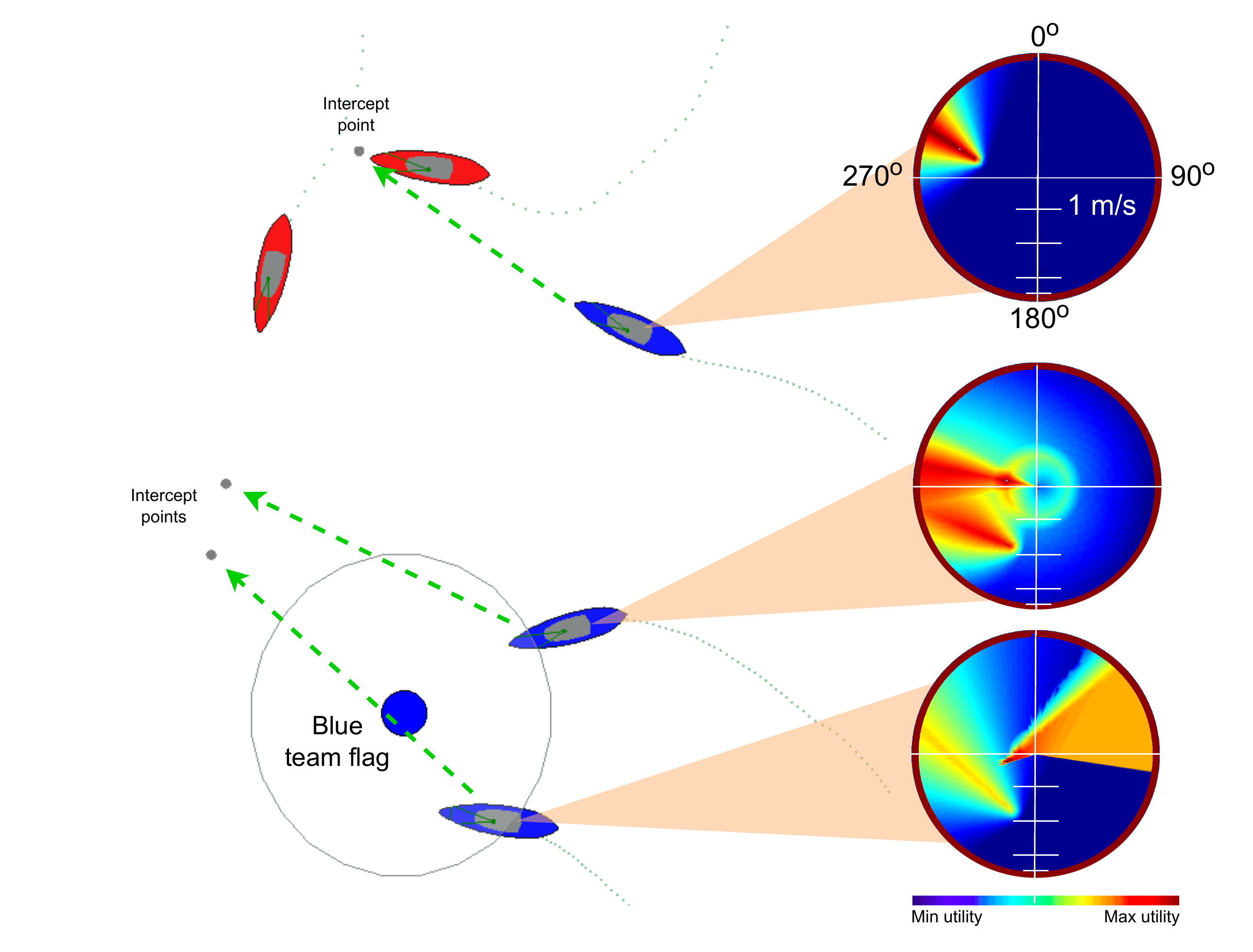}
  \caption{Example of distributed team assignment and non-convex utility functions generated by the combination of intercept and avoid collision behaviors.  The three blue-team vehicles collectively choose the option to defend, and the intercept behavior uses the Hungarian Algorithm to determine optimal assignments as shown by the intercept points. In the case of the lower two blue vehicles, the optimal individual decision for heading and speed is computed using a combination of the utility functions from the intercept and collision avoidance behaviors. }
  \label{fig:CTF_behaviors}   
\end{figure}

This scenario highlights the benefit of using heterogeneous behaviors over homogeneous behaviors typically found in multi-agent consensus such as those based on symmetric potential field-based attraction and repulsion.  
One example of the multi-objective behavior design is shown in Figure \ref{fig:CTF_behaviors}.

The defensive strategy is implemented using a behavior that adjusts dynamically to the situation. Combinatorial optimization is used to first assign vehicles that are defending to incoming intruders based on distance and difference in heading, then the appropriate course is determined.  When more than one defender is assigned to an intruder the group employs a collaborative tagging strategy. The closest defender will make a course to intercept towards a point that is a short distance in front of the intruder, while subsequent defenders that are farther away will make a course to intercept towards a point that is increasingly further ahead.  An example of the strategy in a simulated game is shown in Figure \ref{fig:CTF_behaviors}.

\subsection{Simulated and Experimental Results}
\begin{figure}[ht]
  \centering
\includegraphics[width = \columnwidth]{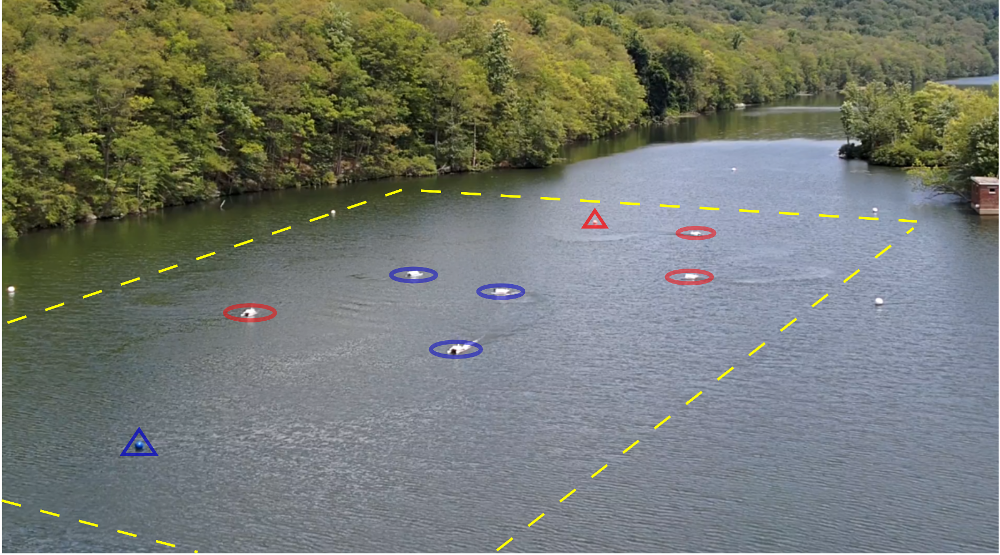}
  \caption{Field competition of 3 vs 3 Aquaticus at US Milliary Academy West Point. Three vehicles on each team (red and blue) play a game of capture the flag.   Approximate location of field is outlined in yellow, flags are marked with triangles, and vehicles are marked with circles. } 
  \label{fig:CTF_WP_Demo}   
\end{figure}

Using this model, a three-agent team competed in a field competition held at the US Military Academy West Point.  
Each team consisted of three Sea Robotics Surveyor M1.8 USVs, and the teams played ten minute games on a field in Lake Popolopen as captured in the image in Figure \ref{fig:CTF_WP_Demo}.  

\begin{figure}[ht]
  \centering
\includegraphics[width = 0.7\columnwidth]{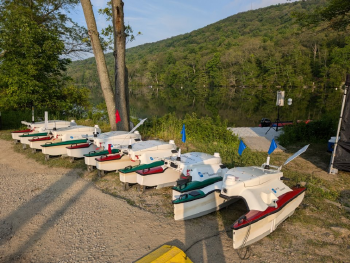}
  \caption{Sea Robotics Surveyor M1.8 USVs used at the US Military Academy West Point.  } 
  \label{fig:CTF_WP_surveyors}   
\end{figure}

We report both simulated an field competition results in Table \ref{tab:CTF_comp_results}.
The opponent for simulation round 1 was the Default strategy, a three vehicle default strategy based on the two vehicle default version reported in \cite{beason2024evaluatingcollaborativeautonomyopposed}.  The Best Rule-Based strategy was a straightforward extension of the winning rule-based strategy reported in \cite{beason2024evaluatingcollaborativeautonomyopposed}.  
The third opponent in simulated competitions was the CBPA model without any notion of group allocation, i.e. the influence parameters where all zero.  This strategy was called ``NO NOD” and offers a perspective of the benefit of NOD in an ablation study. 

The two opponents for the field competition used reinforcement learning-based strategies and were competitive entries in the 2024 Maritime Capture-the-Flag Competition held at the 2024 Autonomous Agents and Multi-Agent Systems (AAMAS) Conference \cite{AAMAS2024CTFComp}.  

\subsection{Discussion}
The strategy developed with the CBPA model performs better on average than both the Default and the Best Rule-Based strategies.   The win rate against the Default strategy is relatively high at 89.0\%, but a more modest 65.0\% against the Best Rule-Based strategy.  These results are encouraging and illustrate the benefits of both the flexible allocation between attacking and defending, and the more sophisticated collaborative defending behavior.   The performance in the ablation study, where this strategy competed against a reduced version of itself, is more interesting.  The win rate is only 50.6\%, with a tie rate of 11.7\%.   These results suggest the behavior design is primarily responsible for the majority of the boost in performance, but there is still a benefit to using NOD method for allocation.  It is possible this benefit from using NOD could grow as the number of team members increases beyond three.

%%%%%%%%%%%%%%%%%%%%%%%%%%%%%%%%%%%%%%%%%%%%%%%%%%%%%%%%%%%%%%%%%%%%%%%%%%%%%%%%%
%\clearpage
\begin{table*}[t]
\fontsize{10.5}{10.5}\selectfont
\centering
\caption{Adaptive Search and Sample + Migration Mission Design (Experiment 3)}
\begin{tabular}{c l l l}
%\toprule
  \textbf{Option Structure }($\bm{f}_{nod}$)  & \textbf{Option}  & \textbf{Option Input} ($\bm{f}_{opt}$)    & \textbf{Behaviors }($\bm{f}_{obj}$) \\
\toprule
 
\raisebox{-0.8\height}[0pt][0pt]{\includegraphics[height=5\normalbaselineskip]{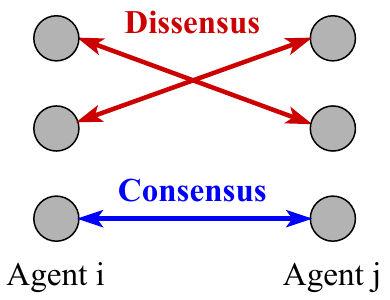}}
  & Search & $f_{search}$  (Section \ref{sec:searching})   & Voronoi, Waypoint (MDP-based) \\
  \cmidrule[\lightrulewidth]{2-2} \cmidrule[\lightrulewidth]{3-4}
  & $f_{sample}$  (Section \ref{sec:sampling})   & Sample & Waypoint (TSP-based) \\
  \cmidrule[\lightrulewidth]{2-2} \cmidrule[\lightrulewidth]{3-4}
  & $f_{migrate}$ (Section \ref{sec:migrating}) & Migrate & MoveToRegion   \\
  &    &    &  \\
\bottomrule
\end{tabular}
\end{table*}

\section{Experiment 3: Adaptive Seek and Sample Scenario}\label{sec:adaptive_sample_demo}
\begin{figure}[ht]
  \centering
\includegraphics[width = \columnwidth]{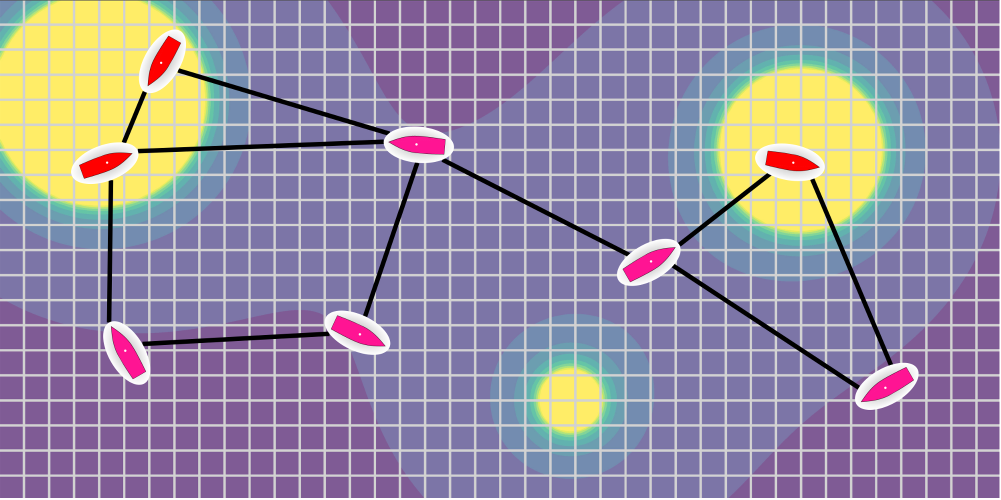}
  \caption{Collaborative adaptive search and sample mission where the goal is to find and sample time-varying hot spots (yellow).  Vehicles are dynamically allocated between cooperative searching (pink) or cooperative sampling options (red).  Network connections are shown in black.  } 
  \label{fig:EEM_mission}
\end{figure}

In this section we consider the common scenario where the population must locate a region of interest, travel there and perform a task such as sample collection.  In just the field of marine robotics, this type of strategy is commonly used for scientific observational studies \cite{mccammon2021JFR}, \cite{flaspohler2019MSS} and military contexts \cite{sanem2008MCMission}.  The type of mission elicits strategies that balance the dual objectives of exploring and exploiting.  
The third objective of migration, or the traveling to another area as a group, is simultaneously considered by the group. Including this third option demonstrates the ability to design more sophisticated group behavior, and is motivated by social organisms that forage cooperatively for food and move between areas together as in a herd or flock.    
This type of mission was introduced in \cite{paine2024ICRAGCID}, and an overview of the this scenario is shown in Figure \ref{fig:EEM_mission}.   The focus of the remainder of this section is on the decentralized allocation between searching (exploring) and sampling (exploiting). We significantly extend the work of Paine and Benjamin in \cite{paine2024ICRAGCID} by: (1) developing a principled method for designing the inputs instead of using heuristically tuned parameters, (2) comparing the improved method against a more capable existing method, and (3) analyzing  the sensitivity to key parameters.  

The class of adaptive seek and sample scenario hinges on $N_{samples}$, the number of samples that need to be collected once an agent detects the region of interest, and their location in space.  At one extreme, when $N_{samples} = 1$ and samples can be collected by the searching vehicle that detected it, then the scenario reduces to one focused solely on joint path-planning to maximize search efficiency, which can be solved using methods such as those reported in \cite{flaspohler2019MSS} and \cite{mccammon2021JFR}.  On the other hand, for $N_{samples} \gg 1$, where a large area must be searched when a detection is received, the scenario reduces to an efficient coverage problem which can be solved using the methods in \cite{Karapetyan2018ICRADubins}, or multi-agent TSP which can be solved using the method in \cite{Luc2009CBBA}.  The most interesting scenario, and the focus of the study herein, is when $N_{samples}$ take a value between these two extremes. In this case the group must balance the objectives of exploring and exploiting. 

We design the system using CBPA framework to achieve three objectives at the group level:
\begin{itemize}
    \item Minimize transit cost (distance) to sample known locations
    \item Maximize expected reward in searching for new locations
    \item Distribute the energy cost among members of the group.  See Section \ref{sec:balance_batt}. 
    \item Be responsive to perceived signal from other USVs to migrate. 
\end{itemize}

\subsection{Adaptive Seek and Sample Formulation}
The adaptive seek and sample problem can be formulated as a minimization of cost over options. 
\begin{align}\label{eq:EEM_cost}
\min_{\bm{z} \in \bar{Z}} J = \min_{\bm{z} \in \bar{Z}} \  J_1 - \eta_1 J_2 + \eta_2 J_3
\end{align}
where $\eta_1, \eta_2$ \st{, $\eta_3$}$\in {\rm I\!R}_+$ are design weights and 
\begin{itemize}
	\item $J_1$ is the total cost (distance) for the group to complete the sampling task(s).  Given a set of known sample locations, $J_1$ is the cost for at least one agent in the population to visit each location, otherwise known as the fleet TSP.   This term is considered in Section \ref{sec:searching}.
	\item $J_2$ is the total expected value for group to explore unknown areas.  More specifically, assuming the area is divided into grids, what is the total expected value of each agent exploring their nearby reachable grids in some finite time horizon?  This term is considered in Section \ref{sec:sampling}.
	\item $J_3$ is the variance in the battery exhaustion among USVs.  This cost is analyzed in detail in Section \ref{sec:balance_batt}. 
\end{itemize}

\subsection{Searching}\label{sec:searching}
Agents balance two objectives when searching: Driving towards the centroid of their Voronoi cell in an effort to disperse across the region, and traveling towards a grid cell within the Voronoi cell with the highest expected value, thus maximizing $J_2$.  The trade off is similar to the one found in Patrolling as described in Section \ref{sec:patrolling}.  Behavior priority weights are used to balance these objectives provided the USV can compute the value the expected value of neighboring grid cells.

The expected value of neighboring grid cells is computed using a Markov Decision Process (MDP) with the maximum value information (MVI) reward introduced in \cite{flaspohler2019MSS}.  This state-of-the-art reward is computed for each grid-cell contained within the agent's Voronoi cell that is reachable from the current position within a specified number of transitions, or tree depth.  The reward is a function of the normalized expected probability that a valid sample location is in that cell, denoted by the $\mathbb{E}[\varphi]$, and the associated uncertainty $\mathbb{V}[\varphi]$.  Due to limited communication between vehicles we assume the following:
\begin{itemize}
    \item Each vehicle makes measurements of $\varphi \in [0,1]$, where a value of $1$ is the detection threshold that triggers a sample request. Each vehicle has a sensor that measures the scalar field with a fixed uncertainty.
    \item Neighboring vehicles will notify if a sample location is determined (they detected $\varphi = 1$), implying that if a neighbor is observed to be within a cell and has not communicated a detection then the measurement is $\varphi = 0.5$ with $\mathbb{V}[\varphi] = (0.1)^2$.
    \item Each vehicle maintains a local grid, and since the phenomena to be sampled is time-varying, the uncertainty of all cells increases to a maximum of $\mathbb{V}[\varphi] = (0.25)^2$ and the value normalizes to $\varphi = 0.5$. These are also the initialization values for each cell in the grid. 
\end{itemize}

The input preference for searching is the estimated marginal improvement in the joint value $J_2$ given the agent chooses to pursue searching.  More concretely, the search-sample allocation is a two-option dissensus case modeled by (\ref{eq:2optionNOD}) where agents with $z_i > 0 \ (< 0)$ opinions choose to search (sample).  The marginal increase in the reward $J_2$ is stated as
\begin{equation}\label{eq:marginal_improv_search_reward}
    f_{search} = \frac{\partial J_2}{ \partial z_i} \approx \bar{J}_{2_i} \big|_{\text{agent i samples}} - \bar{J}_{2_i} \big|_{\text{agent i searches}}.
\end{equation}
where $\bar{J}_{2_i}\big|_{\text{agent i searches}}$ is the local computed value using observations of agent $i$ given that agent $i$ opts to search. 
This approximation of $\frac{\partial J_2}{ \partial z_i}$ is justified because it is impossible to compute the true marginal improvement without global information, and the following process uses state-of-the-art optimization methods with the information that is available. 

In both cases, the ego agent computes the expected cumulative  reward from all neighbors given local knowledge as shown in Figure \ref{fig:EEM_mission}. 
\begin{figure}[ht]
  \centering
\includegraphics[width = \columnwidth]{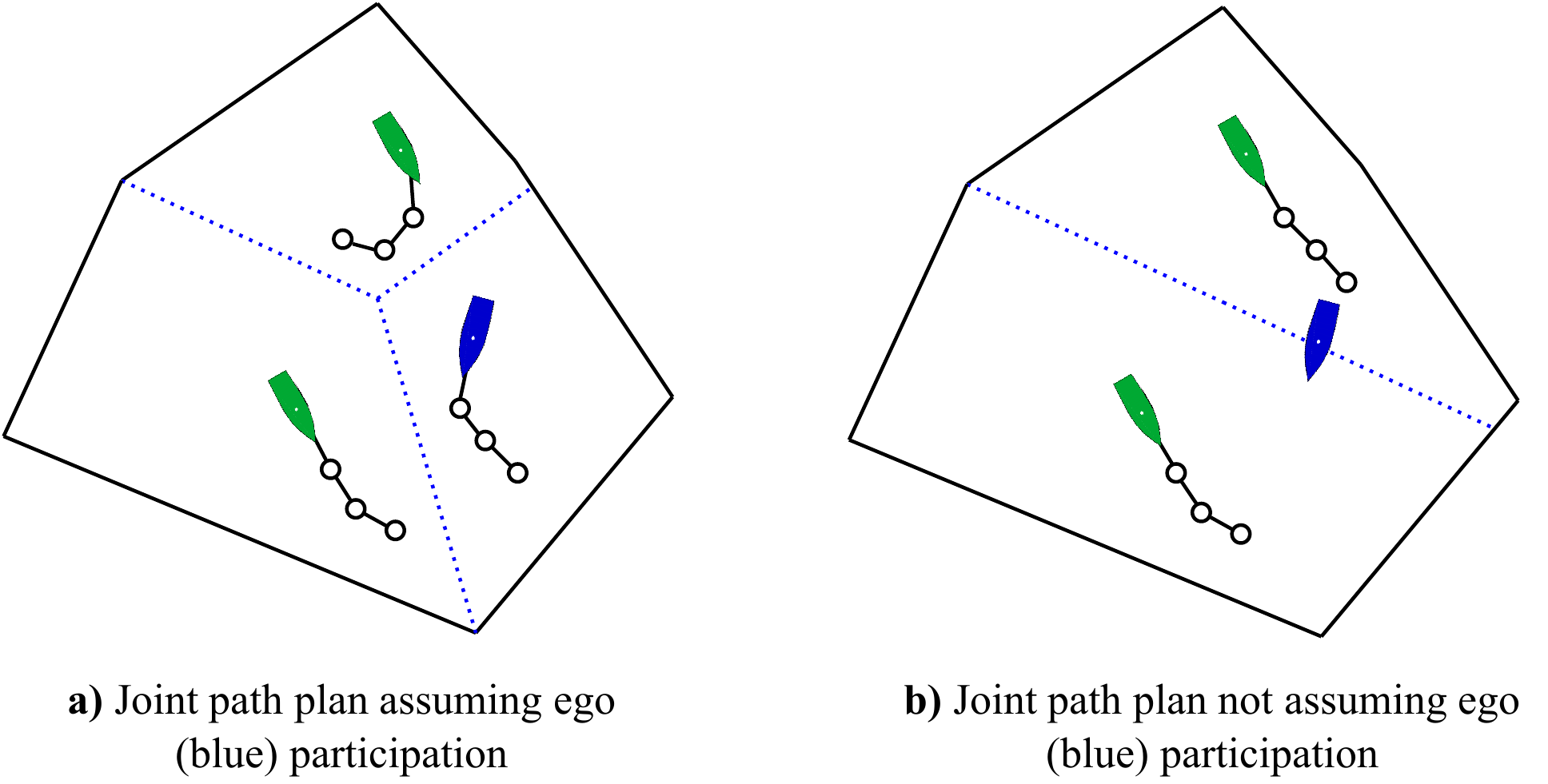}
  \caption{Method to compute the estimate of the marginal improvement in reward (\ref{eq:marginal_improv_search_reward}) by selecting the option to explore.} 
  \label{fig:EEM_search_calc}   
\end{figure}

\subsection{Sampling}\label{sec:sampling}
Sampling vehicles seek to minimize $J_1$, the global cost to travel and collect samples at all known locations.
Each agent estimates the marginal decrease in $J_1$, i.e., 
\begin{equation}\label{eq:marginal_improv_sample_cost}
    f_{sample} = \frac{\partial J_1}{ \partial z_i} \approx \bar{J}_{1_i} \big|_{\text{agent i searches}} - \bar{J}_{1_i} \big|_{\text{agent i samples}},
\end{equation}
where $\bar{J}_{1_i}\big|_{\text{agent i searches}}$ is the local computed cost using observations of agent $i$ given that agent $i$ opts to sample.  

If the locations of both the samples and neighboring vehicles are known, then computing the cost terms in (\ref{eq:marginal_improv_sample_cost}) is a complex, but well known variant of a TSP.  The Hungarian algorithm is used to iteratively assign sampling agents to sample locations, providing an approximation of the cost to complete all samples.  The algorithm is completed twice, once assuming ego participation in sampling, and again without ego participation.

\subsection{Migrating}\label{sec:migrating}

The input for migration is a relatively large positive constant value if a migration signal is detected and otherwise remains at zero.  Migration behavior takes advantage of the tunable attention mechanism introduced in \cite{Bizyaeva2023OD_TAC}.  By design the attention \eqref{eq:atten_dynam} is elevated from a nominally low level to a higher level when a neighboring vehicle has a strong opinion, which is the case when that agent has a large input from a migration signal.  The resulting collective behavior is latching, a feature further explored in the theoretical analysis of opinion cascades \cite{bizyaeva2021Cascades}.

\subsection{Simulated Results}
\begin{figure}
    \centering
    \includegraphics[width=0.9\columnwidth]{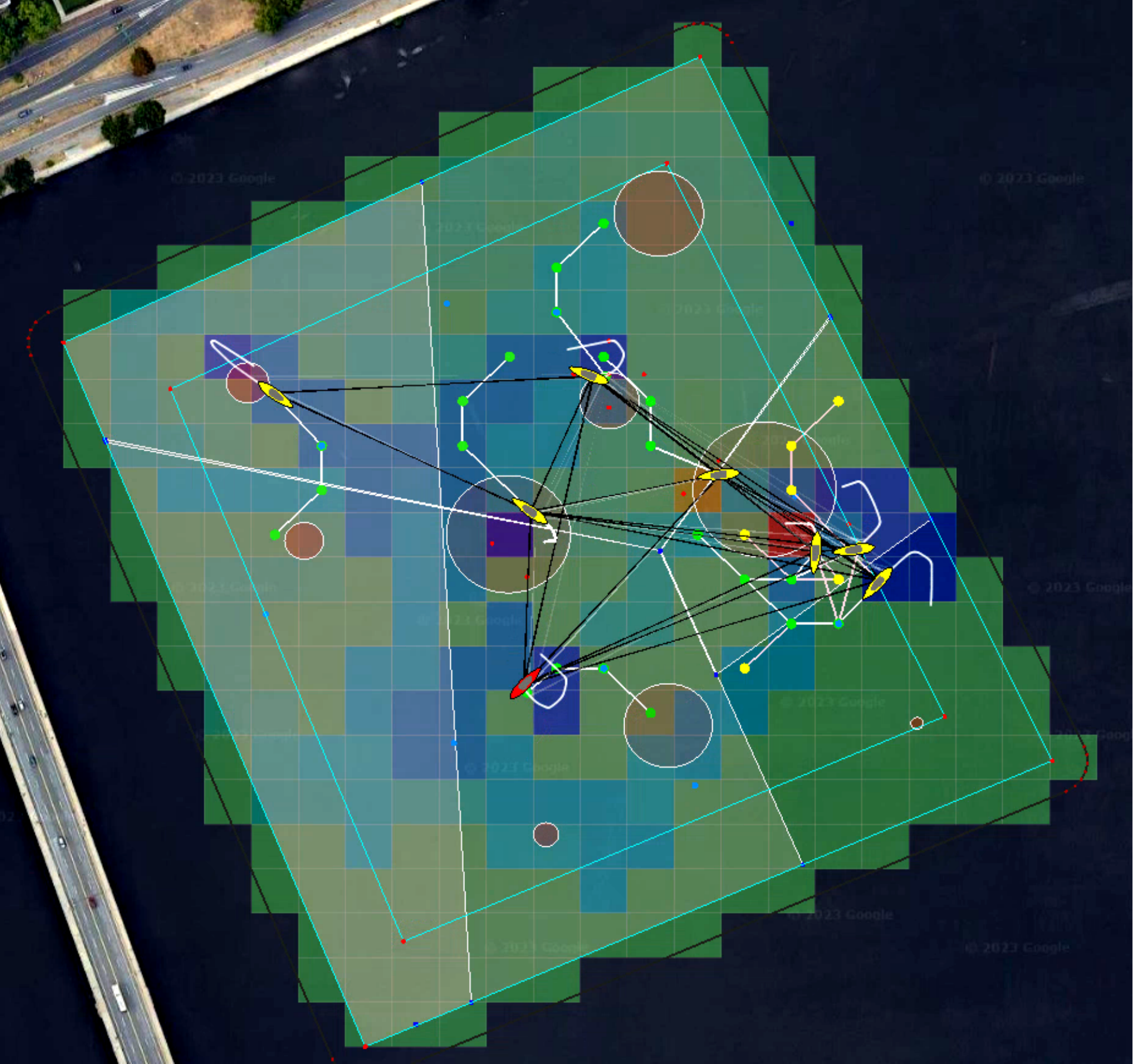}
    \caption{State of system during the adaptive seek and sample mission. Unknown regions of interest are in red, and the grid values range from blue (no interest found) to red (found region of interest).  Sampling location are the small red circles. Black lines indicate network connections between vehicles.  Four vehicles are exploring within their Voronoi cell and their computed path with green vertices is directed towards the unknown sample locations as designed. Four vehicles are allocating samples via a distributed solution to the TSP. Vehicle paths with yellow vertices are computed by neighboring vehicles as the most probable path that vehicle would select if the neighbor selected to explore, given the neighbor's estimate of the grid. }
    \label{fig:EEM_Sim}
  %  \vspace{-5mm}
\end{figure}

\begin{figure}
    \centering
    \includegraphics[width=1\columnwidth]{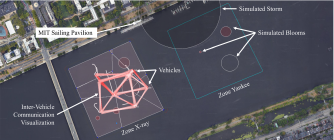}
    \caption{State of the MRS and simulated environment approximately 50 minutes into a two hour mission on the Charles River \cite{paine2024ICRAGCID}.  Simulated blooms appeared randomly within both zones X-ray and Yankee, and grew larger with time.  A randomly generated storm periodically passed over the regions. At this time during the mission, 8 vehicles were searching and sampling in Zone X-ray.  The communication range was artificially limited to 160 meters, and the restricted inter-vehicle communication is visualized in red. Both zones measured 300 meters by 350 meters.   }
    \label{fig:BloomStormMission}
\end{figure}

Simulation was used to validate the approach and also show how to further optimize performance for a specific instance of this scenario with a unique set of parameters, such as the frequency of sample areas, detection probability, etc.  An overview of the mission introduced by Paine et al. \cite{paine2024ICRAGCID} was used in both simulation and field demonstrations is shown in Figure \ref{fig:BloomStormMission}. Simulated regions to be sampled start from a randomly generated location within the zone and grow outward for ten minutes.  A team of eight Heron USVs use the approach described herein to minimize (\ref{eq:EEM_cost}) and the elements of this approach can be seen in Figure \ref{fig:EEM_Sim}.

\begin{figure}
    \centering
    \includegraphics[width=1.0\columnwidth]{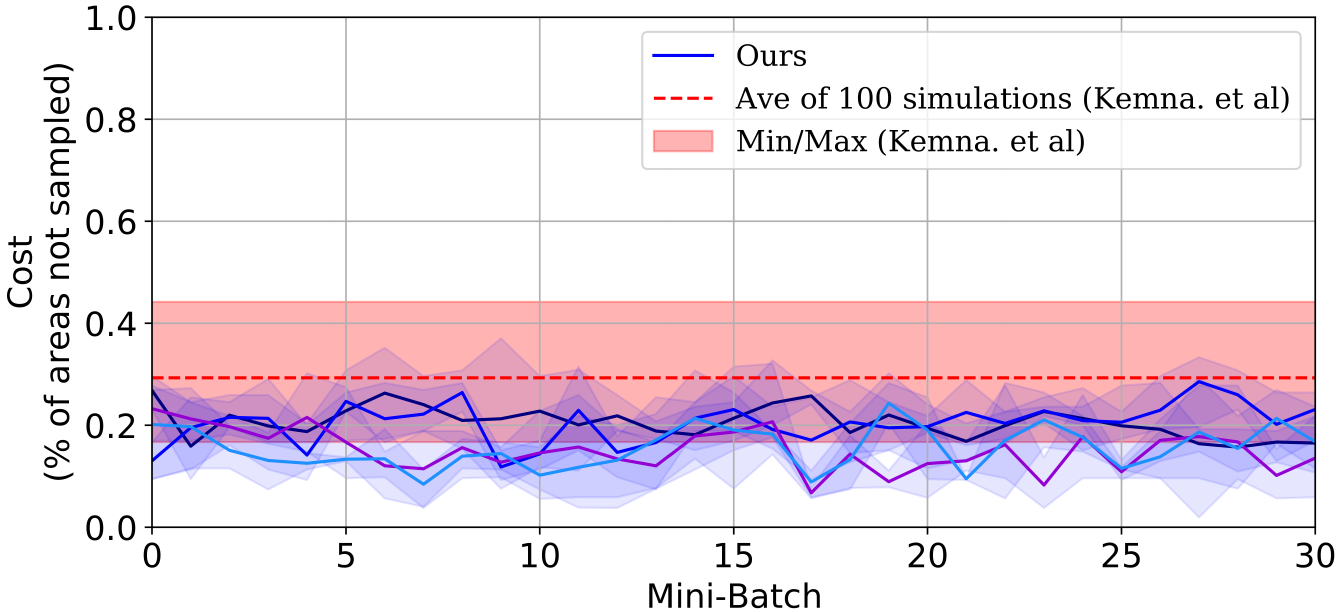}
    \caption{Results using VPG for $\eta_{1}$, $\eta_{3}$, and $\eta_{4}$ starting from different initial values. The results from \cite{Kemna2017ICRA} give an ablation study where there is no collective option allocation between sampling and search.}
    \label{fig:EEM_Sim_Loss}
\end{figure}

Using this simulated environment we investigated the sensitivity of the performance with respect to key parameters related to the explore-exploit tradeoff using vanilla policy gradient (VPG) iteration \cite{Albrecht2024marl-book}.  The results are shown in Figure \ref{fig:EEM_Sim_Loss}. 
Although we describe the process for computing a proxy for the marginal improvement terms $\frac{\partial J_1}{ \partial z_i}$ and $\frac{\partial J_2}{ \partial z_i}$, the question that remains is how to select the scalar $\eta_{1}$ that enters into the cost function and thus appear in the approximation of the gradient term $\bm{f}_{opt}$.  Since the marginal improvement terms are approximations, a bias term $\eta_{3}$ was included in addition to a scalar factor $\eta_{4}$ applied to all terms in $\bm{f}_{opt}$. Only the key parameters were included in this VPG iteration study to reduce the dimension of the search space, and for this study only the input was 
\begin{equation}
    \bm{f}_{opt} = \eta_4 \cdot \big(\frac{\partial J_1}{ \partial z_i} - \eta_1\frac{\partial J_2}{ \partial z_i} + \eta_{3} \big)
\end{equation}

The metric of performance in this mission is the percentage of areas that were not sampled. The results in Figure \ref{fig:EEM_Sim_Loss} suggest that the CBPA model outperforms another approach reported in \cite{Kemna2017ICRA}, a method with the same Voronoi-based search algorithm, but without the dynamic allocation of USV into search and sample sub-teams.  Moreover, the results indicate that the CBPA model performance is not highly dependent on selecting optimal parameter values for $\eta_{1}$, $\eta_{3}$, and $\eta_{4}$. 
The ranges were $\eta_{1} \in [67.8, 123.7]$, $\eta_{3}\in [67.7, 123.7]$, and $\eta_{4} \in [7.9, 55.7]$. 
There is only sparse evidence that further performance gains can be realized via policy iteration.

\subsection{Experimental Results}
We report an analysis of a two hour night-time field experiment on the Charles River reported in \cite{paine2024ICRAGCID} from the perspective distributed teaming.  Using the CBPA approach with heuristically tuned $f_{sample}$ and $f_{search}$ functions, all 8 USVs dynamically changed teams, locating sampling areas and allocating who should sample them.  All vehicles selected to start in the eastern zone and all migrated to the western zone when one vehicle detected a storm.   

Overall, the performance of the heuristically tuned algorithm in the field matched the simulated results. Each vehicle transitioned from sampling to searching at least once, and no vehicle had a network degree equal to seven throughout the entire mission, indicating no vehicle acted as a central network hub. Other conclusions of this field test were that it is possible to run this model on relatively low-cost computers such as a Raspberry Pi 4, and communicate adequately with commercially available RF antennas.  Finally, as reported in \cite{paine2024ICRAGCID} one vehicle failed during the mission and was later revived to rejoin the group.  The rest of the working Herons were able to continue without any intervention from the human operators, further demonstrating the robustness of the CBPA approach in field conditions.

%%%%%%%%%%%%%%%%%%%%%%%%%%%%%%%%%%%%%%%%%%%
%% Conclusion
\section{Conclusion}\label{sec:conc}
The census-based population autonomy model described in this paper builds upon foundational algorithms for distributed autonomy to realize new types of collective decision-making that include heterogeneous behaviors with non-convex cost functions and optimization of cost functions that are not fully observable. The model maintains scalability to large group sizes.  While the model is general, there are immediate and specific applications, a few of which are described in detail in this paper.   Using three categorically different missions we detail the design of key functions and behaviors.  The results were compared against either a centralized approach, or at least one leading alternative method.  

This work motivates new and exciting questions for future research.  First, an exploration of the validity of the mean field assumption, which underpins much of the recent research in the field of multi-agent autonomy including the approach introduced here, is needed.  Another interesting question is whether the second-order optimization approach that was introduced in this paper is a valid model of the collective decision-making process observed in groups of biological organisms.

\bibliographystyle{IEEEtran}
\bibliography{IEEEabrv,refs_ICRA24,refs_Notes,refs_others}

\begin{IEEEbiography}[{\includegraphics[width=1in,height=1.25in,clip,keepaspectratio]{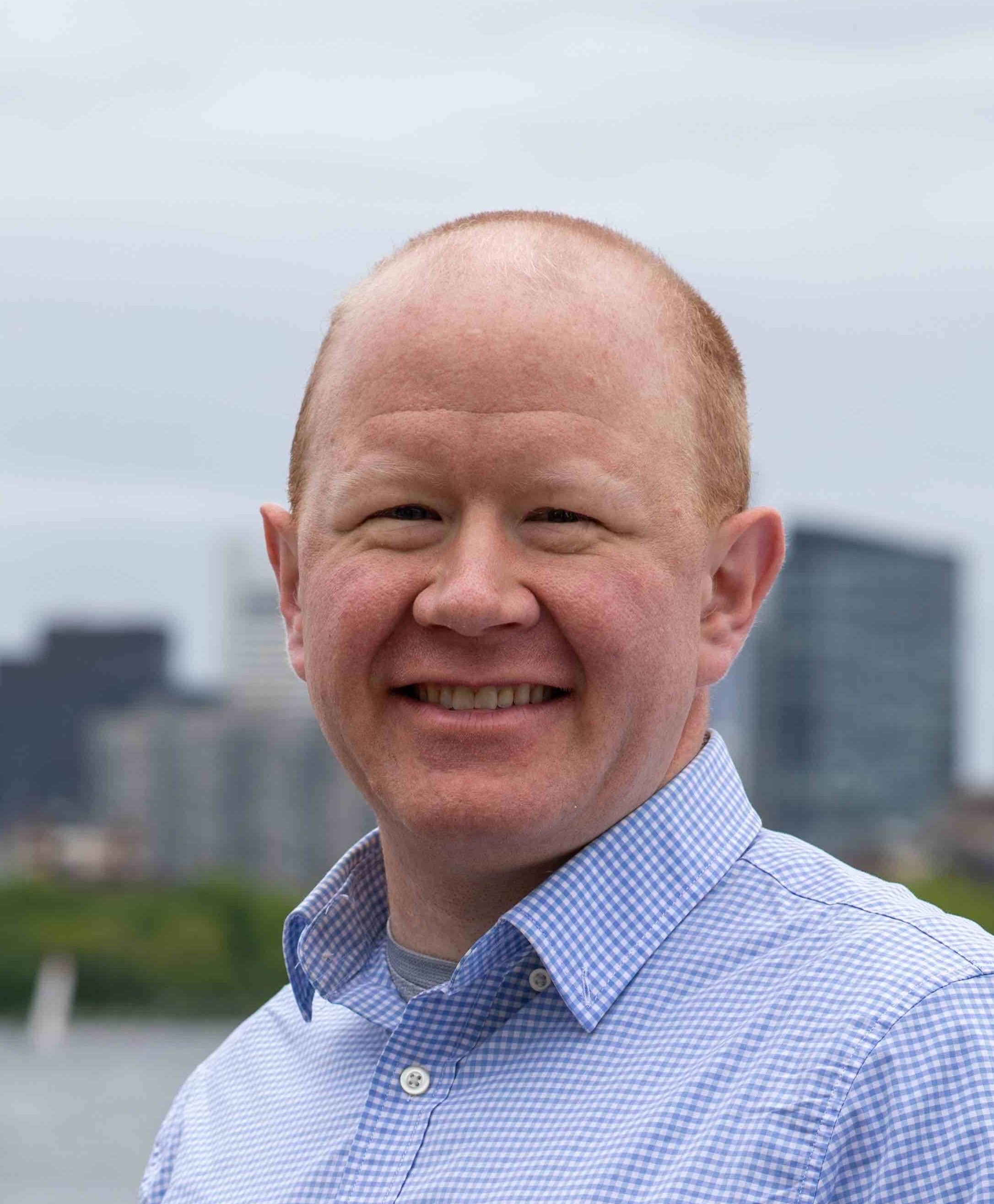}}]{Tyler Paine}
received the B.S. in mechanical engineering from Montana State University, Bozeman, MT, USA, in 2011, the M.S. in mechanical engineering from Johns Hopkins University, Baltimore, MD, USA, in 2018, and the Ph.D. in mechanical and oceanographic engineering from the Massachusetts Institute of Technology, Cambridge, MA, USA, and the Woods Hole Oceanographic Institution Joint Program. 
From 2011-2020 he was an engineer with NAVSEA Division Keyport.  He is currently a research scientist with the Department of Mechanical Engineering at Massachusetts Institute of Technology.  His research interests include the design, analysis, and deployment of multi-agent marine robotic systems. 
\end{IEEEbiography}
\begin{IEEEbiography}[{\includegraphics[width=1in,height=1.25in,clip,keepaspectratio]{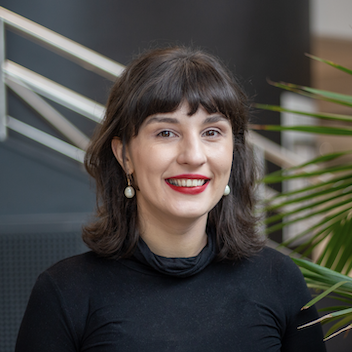}}]{Anastasia Bizyaeva} received the B.A. degree in Physics with a minor in Mechanical Engineering from University of California, Berkeley, Berkeley, CA, USA, in 2016, and the M.A. and Ph.D. degrees in Mechanical and Aerospace Engineering from Princeton University, Princeton, NJ, USA in 2019 and 2022, respectively. She is currently an assistant professor in the Sibley School of Mechanical and Aerospace Engineering at Cornell University where she leads the Control and Computation for Complex Systems Group. Previously she was a postdoctoral scholar with the NSF AI Institute in Dynamic Systems at the University of Washington. Her research explores mathematical connections between flexible and collectively intelligent behavior in biological and socio-cognitive systems, and the design of autonomous behaviors in engineered teams. She is a recipient of the 2025 IEEE Control Systems Society George S. Axelby Outstanding Paper Award.
\end{IEEEbiography}

\begin{IEEEbiography}[{\includegraphics[width=1in,height=1.25in,clip,keepaspectratio]{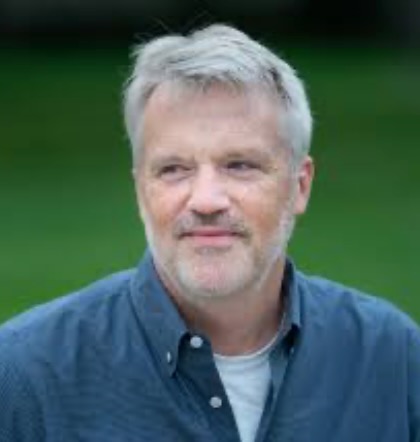}}]{Michael Benjamin}
is a Principal Research Scientist at MIT in the Department of Mechanical Engineering. His research focus is on autonomy algorithms and software for unmanned marine vehicles. In 2005 he founded an open source project named moos-ivp.org comprising dozens of marine autonomy applications including the IvP Helm for autonomous decision making, and COLREGS autonomy on unmanned surface vessels. He received the BS and MS degrees in computer science and cognitive science from Rensselaer, and the MS and PhD degrees in computer science from Brown University. Prior to coming to MIT, he was a research scientist at the Naval Undersea Warfare Center in Newport RI.
\end{IEEEbiography}

\end{document}